\documentclass[11pt]{article}

\usepackage[preprint]{acl}
\usepackage{times}
\usepackage{latexsym}
\usepackage[T1]{fontenc}
\usepackage[utf8]{inputenc}
\usepackage{microtype}
\usepackage{inconsolata}
\usepackage{graphicx}
\usepackage{booktabs}
\usepackage{array,balance}
\usepackage{amsmath,subcaption}
\usepackage{multirow}
\usepackage{enumitem}
\usepackage{listings}
\usepackage[most]{tcolorbox}
\graphicspath{{figures/}}

\lstdefinestyle{promptstyle}{
  basicstyle=\ttfamily\scriptsize,
  breaklines=true,
  breakatwhitespace=false,
  columns=fullflexible,
  keepspaces=true,
  showstringspaces=false,
  upquote=true
}

\newtcblisting{promptboxside}[1]{
  enhanced,
  colback=black!2,
  colframe=black!35,
  boxrule=0.45pt,
  arc=1.2pt,
  left=4pt,
  right=4pt,
  top=4pt,
  bottom=4pt,
  title={#1},
  fonttitle=\bfseries\small,
  coltitle=black,
  colbacktitle=black!8,
  listing only,
  listing options={style=promptstyle}
}

\newcommand{\dataset}{IEMO-MECP}
\newcommand{\cause}{\texttt{emo-cause}}
\newcommand{\context}{\texttt{emo-context}}
\newcommand{\nonpair}{\texttt{non-pair}}

\newcommand{\mthreehg}{M\textsuperscript{3}HG}
\newcommand{\mecpetwostep}{MECPE-2step}
\newcommand{\mecpetwosteptable}{\shortstack{MECPE\\[-1pt]-2step}}

\title{\emph{They Are Not the Same:} \\ Direct-Cause Extraction Is Not Grounded Emotion Explanation}

\author{
Zhuangzhuang Pan\textsuperscript{1}\thanks{equal contribution; authors are listed alphabetically by surname.}\,\,
Yan Xia\textsuperscript{1,2}\footnotemark[1]\,\,
Chee Seng Chan\textsuperscript{1,3}\thanks{Corresponding author (cs.chan@um.edu.my).} \\
\textsuperscript{1}Universiti Malaya, Malaysia \\
\textsuperscript{2}Suzhou University of Technology, China \\
\textsuperscript{3}VinUniversity, Vietnam \\
\texttt{\{23078403; 23072126\}@siswa.um.edu.my, { cs.chan}@um.edu.my}
}

\begin{document}
\maketitle
\begin{abstract}
Emotion-Cause Pair Extraction (ECPE) was introduced to explain why an emotion occurs, but this goal is now often reduced to binary pair/non-pair prediction. This proxy is useful for direct-cause extraction, yet easy to over-read as evidence grounded emotion explanation. We show that this interpretation is only partially valid. In \dataset{}, 90.9\% of original positives remain \cause{} and 95.0\% of original negatives remain \nonpair{}, confirming that the binary ECPE task is largely preserved. The problem is that direct triggers alone do not constitute a grounded explanation. \context{}, an utterance that helps interpret a target emotion without directly causing it, appears on both sides of the original boundary and is enriched near binary uncertainty, showing that the binary boundary has no stable place for such discourse evidence. Across evaluated ECPE models, direct triggers are recovered more reliably than contextual support. Under shortcut pressure, this imbalance becomes consequential. Binary-trained models assign higher pair scores to nearby lexically similar \nonpair{} candidates than to evidence supported but structurally harder \cause{} and \context{} pairs. Thus, pair scores can reward convenient attributions over grounded explanations. High binary ECPE performance indicates that a model can identify direct triggers; it does not indicate that the model has explained the emotion. Code is publicly available at \url{https://github.com/panzhzh/ECPExsame}.
\end{abstract}

\section{Introduction}
Emotion-Cause Pair Extraction (ECPE) was introduced to explain why an emotion occurs by recovering emotion-cause pairs from text \citep{xia-ding-2019-emotion}. The task has since moved from documents to conversations, where causes may depend on speaker turns, interaction history, and dialogue structure \citep{poria2021recognizing,li2023ecpec,jeong-bak-2023-conversational}, and further to multimodal conversations that combine textual, acoustic, and visual cues \citep{wang2023multimodal,wang-etal-2024-semeval,li2025multimodal}. 

\begin{figure}[t]
  \centering
  \includegraphics[width=0.95\linewidth]{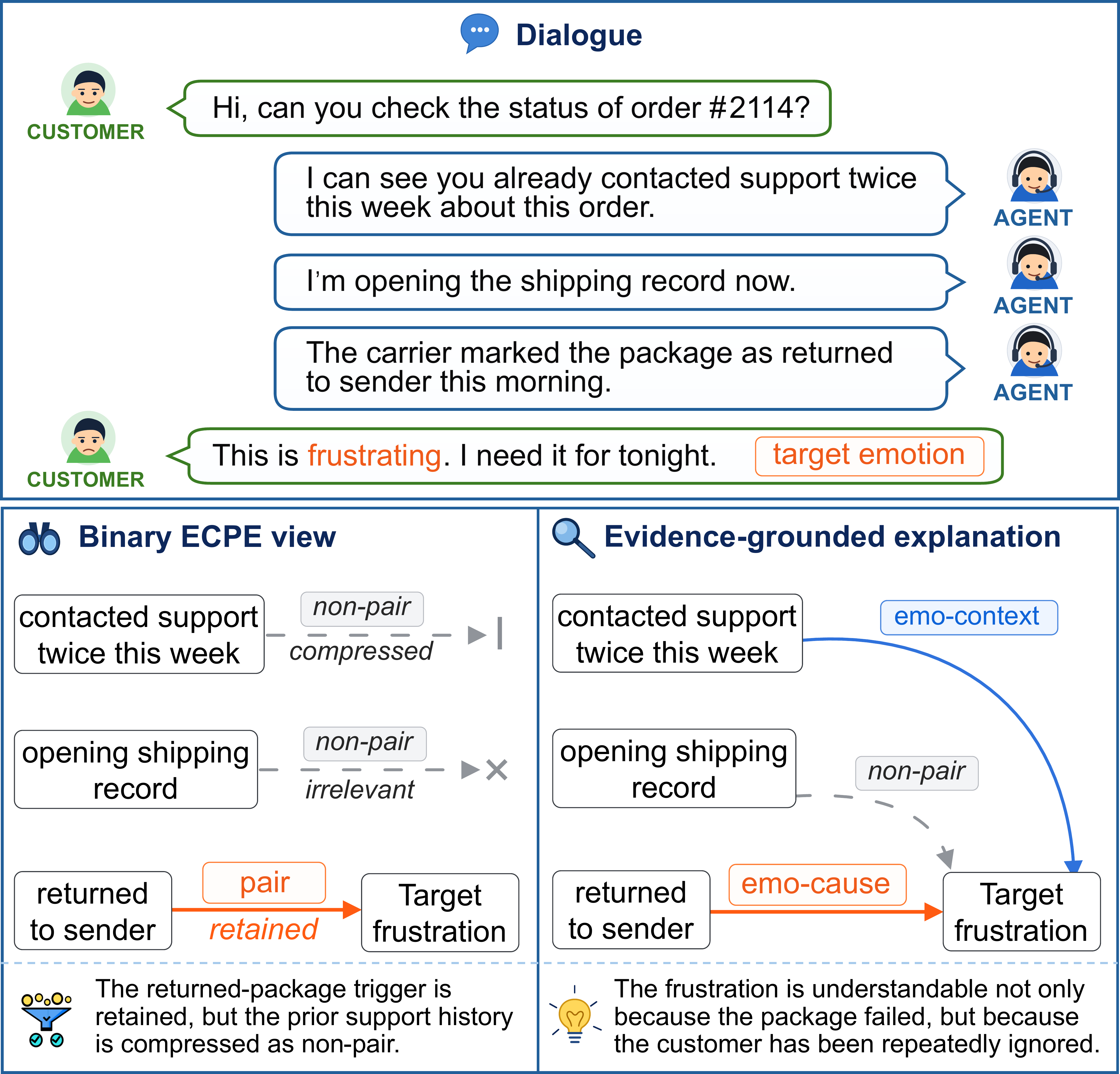}
  \caption{Boundary-compression example. Binary ECPE can retain the direct trigger while compressing discourse evidence needed to understand why the emotion makes sense. The validity risk is not missing the trigger; it is treating trigger extraction as explanation.}
  \label{fig:intro-boundary}
  \vspace{-1em}
\end{figure}

Despite these advances, progress on this explanation oriented task is commonly evaluated through a narrower proxy, namely binary pair/non-pair prediction. This proxy is useful for direct-cause extraction, but it changes the measurement target. A model can succeed at deciding whether an utterance should be counted as a cause pair while still failing to preserve the discourse evidence that makes the emotion understandable in context. The resulting risk is not simply lower performance on hard cases. It is an \emph{\textbf{evaluation validity failure}}, where pair scores may certify trigger extraction while being interpreted as evidence grounded explanation. Prior work has shown that context, position, and commonsense affect emotion-cause reasoning \citep{li-etal-2018-co,yan-etal-2021-position,zhao-etal-2023-knowledge}; our concern is whether the binary pair/non-pair label space can represent such evidence when it is explanatory but not a direct trigger.

Figure~\ref{fig:intro-boundary} illustrates this gap. The prior contact utterance is not a weaker form of the returned package cause. It does not trigger the final frustration, but changes how that frustration is understood. The target utterance reads as part of an ongoing unresolved service problem. This is the \context{} role, contextual support that helps interpret a target emotion without directly causing it. The question is therefore not whether binary ECPE is useful, but what its pair scores can justify. A high pair score may support a direct trigger decision, but should not be read as evidence that the model has recovered an evidence grounded explanation of the emotion. At the label-space level, this is a projection problem, where binary ECPE must place direct causes, contextual support, and unsupported candidates into only two labels. As a result, contextual support is either over-causalized as \textit{pair} or discarded as \textit{non-pair}, and neither choice preserves the full explanation structure.

We study this problem with \dataset{} as a diagnostic audit of the inherited ECPE candidate space, not as a replacement for existing benchmarks. The audit keeps the original target-candidate pairs and source-binary labels, then assigns each pair one of three roles, \cause{}, \context{}, or \nonpair{}. This gives a binary view and a role-aware view of the same candidate set, aligned pair by pair. This alignment matters because disagreements between the two views can be analyzed as evidence about what the binary boundary compresses, not as artifacts of different candidate construction. Our goal is to identify where binary pair scores remain valid, where their explanation claims stop, and why this boundary matters for evaluating emotion explanation systems.

Our contributions are threefold.
\begin{itemize}
    \item We frame binary ECPE scoring as an evaluation validity problem, showing that pair scores support direct-cause decisions but not evidence grounded emotion explanation.
    \item We introduce \dataset{} as a diagnostic audit of the inherited ECPE candidate space, where \context{} marks non-triggering discourse evidence compressed by the binary boundary.
    \item We evaluate ECPE models with role-wise diagnostics and shortcut stress tests, showing that \context{} often collapses into \nonpair{} and unsupported local candidates can outrank harder evidence supported pairs.
\end{itemize}

\section{Evaluation-Validity Framing}
\label{sec:framework}

\subsection{Pair Scores and Explanation Claims}

We formalize the gap between binary pair decisions and explanation claims. Let $D=\{u_1,\cdots,u_n\}$ be a dialogue and let $\mathcal{E}(D)$ denote its non-neutral target emotion turns. Following conversational ECPE, the candidate space is lower triangular:
\begin{equation}
\mathcal{P}(D)=\{(i,j): i\in\mathcal{E}(D),\,1\leq j\leq i\},
\end{equation}
where $u_i$ is the target emotion utterance and $u_j$ is either the target utterance itself or an earlier candidate. Standard binary ECPE assigns $y^{\mathrm{bin}}_{ij}\in\{0,1\}$, indicating whether $u_j$ is treated as a cause of the emotion in $u_i$. We use pair score to denote the confidence assigned to $y^{\mathrm{bin}}_{ij}=1$. This score supports a binary pair decision, namely whether $u_j$ should be counted as a cause of the emotion in $u_i$.

The decision is appropriate when the evaluation target is direct-cause extraction. The validity problem arises when the same score is used as evidence that a model explains why the target emotion occurred. Contextual cues can shape how a target emotion is interpreted without being the direct trigger. Pair-level ECPE evaluation therefore needs a distinction between direct triggers and non-triggering discourse evidence.

\begin{table*}[!t]
  \centering
  \caption{Pair-role criteria for the diagnostic audit.}
  \label{tab:role-semantics-main}
  \scriptsize
  \setlength{\tabcolsep}{3pt}
  \resizebox{\linewidth}{!}{%
  \begin{tabular}{p{0.10\linewidth}p{0.30\linewidth}p{0.32\linewidth}p{0.20\linewidth}}
    \toprule
    Role & Positive criterion & Exclusion criterion & Frequent boundary error \\
    \midrule
    \cause{} &
    Direct trigger, event, or appraisal for the target emotion. Self-pairs are valid when the target utterance contains this evidence. &
    Not for background or contextual support alone. &
    Over-causalizing background evidence. \\
    \addlinespace[1pt]
    \context{} &
    Emotionally relevant but non-triggering; removal weakens target-emotion interpretation. &
    Not for weak topical similarity, proximity, or generic prior dialogue. &
    Treating context as weak cause or soft \nonpair{}. \\
    \addlinespace[1pt]
    \nonpair{} &
    Insufficient evidence for direct causality or contextual emotional relevance. &
    Not when the candidate supplies background, conflict setup, relationship state, or situational evidence. &
    Discarding useful non-triggering discourse evidence. \\
    \bottomrule
  \end{tabular}
  }
  \vspace{-1.0em}
\end{table*}

\subsection{Defining Pair Roles}

We assign each candidate pair a role $y^{\mathrm{role}}_{ij}\in\mathcal{R},$ where $\mathcal{R}=\{c,x,n\}$, such that $c$, $x$, and $n$ correspond to \cause{}, \context{}, and \nonpair{}. These roles are not confidence levels. They describe different evidential relations between a candidate utterance and a target emotion. Let $D_{ij}$ denote observable direct-cause evidence, where the candidate supplies the direct trigger, event, or appraisal for the target emotion. Let $S_{ij}$ denote observable discourse evidence that helps interpret the target emotion without directly triggering it. The intended role hierarchy is
\begin{equation}
y^{\mathrm{role}}_{ij}=
\begin{cases}
c, & D_{ij},\\
x, & \neg D_{ij}\wedge S_{ij},\\
n, & \text{otherwise}.
\end{cases}
\end{equation}
If a candidate both supplies background and directly grounds the target emotion, it is labeled \cause{}. The \context{} role is reserved for evidence that supports interpretation without serving as the direct trigger.

Table~\ref{tab:role-semantics-main} turns this hierarchy into audit criteria. The key boundary is \context{}. A \context{} pair must make the target emotion easier to interpret when included and harder to interpret when removed. It may involve background events, conflict setup, relationship state, discourse bridging, prior speaker state, situational constraints, or alternative plausible causes that shape appraisal. Mere proximity or topical similarity is not enough. This keeps \context{} separate from both direct causes and ordinary \nonpair{} candidates.

\paragraph{Binary projection loses one distinction.}
The role space contains three evidential relations, \(\mathcal{R}=\{\cause{},\context{},\nonpair{}\}\), while binary ECPE observes only \(y^{\mathrm{bin}}\in\{0,1\}\). Any binary evaluation therefore applies a projection $\pi:\mathcal{R}\rightarrow\{0,1\}.$ For direct-cause extraction, the natural projection satisfies \(\pi(\cause{})=1\) and \(\pi(\nonpair{})=0\). The role \(\context{}\) has no projection value that preserves both dimensions relevant to explanation. It is non-causal, but still explanatorily relevant. If \(\pi(\context{})=1\), contextual support is over-causalized as a pair; if \(\pi(\context{})=0\), it is treated as irrelevant. Thus, binary scoring can preserve the direct-cause decision while losing the distinction between unsupported candidates and non-triggering explanatory evidence. See Appendix~\ref{app:formal-compression} for completeness.

\subsection{Diagnostic Predictions}

To make the validity question testable, we turn the role distinction into checks on audited labels and model scores, yielding five diagnostic predictions.

\textbf{P1. Direct-cause preservation.}
If the role-aware view refines rather than replaces the binary task, original positives should largely remain \cause{}, and original negatives should largely remain \nonpair{}.

\textbf{P2. Boundary crossing.}
If the binary boundary compresses contextual support, \context{} should appear among both original positives and original negatives.

\textbf{P3. Uncertainty enrichment.}
If \context{} is compressed by the binary boundary, it should be enriched near binary uncertainty rather than binary-score extremes.

\textbf{P4. Context-channel weakness.}
If the binary objective lacks a context channel, gold \context{} pairs should score closer to \nonpair{} than to \cause{}.

\textbf{P5. Shortcut-evidence conflict.}
If pair scores partly reflect shortcut-compatible cues, unsupported local candidates may outrank evidence supported but structurally harder pairs.

\section{Role Audit and Boundary Diagnostics}
\label{sec:data-protocol}

\subsection{Role-Audited Candidate Space}
\dataset{} is a role-audited extension of ConvECPE \citep{li2023ecpec}, an IEMOCAP-based conversational ECPE dataset \citep{busso2008iemocap}. IEMOCAP provides the underlying multimodal dialogues and emotion annotations, while ConvECPE adds emotion-cause annotations. We keep the inherited target candidate pairs and source-binary labels, and assign each candidate pair the three-role label defined in Section~\ref{sec:framework}.

Role auditing follows the controlled design used by the diagnostics. The inherited source-binary labels are retained as metadata, while role labels are assigned under the criteria in Table~\ref{tab:role-semantics-main}. An LLM provides initial role proposals, which are then reviewed for dialogue-level consistency, refined in high-risk boundary regions, and checked at the final file level. High-risk boundary transitions are additionally reviewed by human auditors. A blind re-annotation audit on 330 boundary-focused judgments yields 0.821 pairwise agreement, mean Cohen's $\kappa$ of 0.732, and Krippendorff's $\alpha$ of 0.732; full provenance, judging, and draft-to-final diagnostics are in Appendices~\ref{app:human-validation}-\ref{app:llm-human-change}.

The audited candidate space contains 86,075 pairs, with 8,635 \cause{}, 4,346 \context{}, and 73,094 \nonpair{} pairs. Full split statistics are reported in Appendix~\ref{app:split-statistics}. The labeling prompt and refinement rules are reported in Appendix~\ref{app:prompts}.

\subsection{Source-Binary Transitions}
\label{sec:source-binary-transitions}
\begin{figure}[t]
  \centering
  \includegraphics[width=0.95\linewidth]{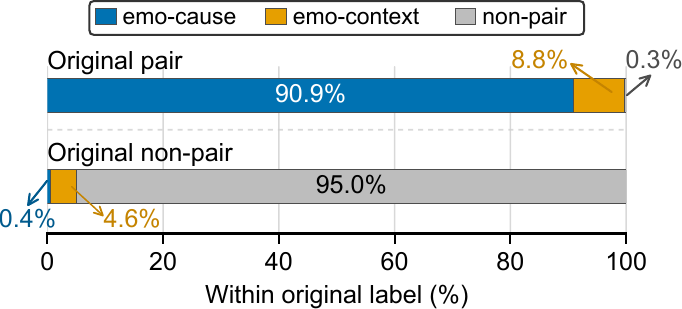}
\caption{Source-binary transitions. The binary task is largely preserved, but \context{} crosses both sides of the inherited pair/non-pair boundary.}
  \label{fig:source-binary-transitions}
\end{figure}

Figure~\ref{fig:source-binary-transitions} compares source-binary and final pair-role labels over the same candidate space. For P1, the large preserved blocks show that the audit largely retains the original direct-cause abstraction with 90.9\% of original positives remain \cause{} and 95.0\% of original negatives remain \nonpair{}. Thus, the boundary problem is not an artifact of replacing ECPE with a different task.

For P2, the smaller \context{} blocks carry the validity signal. Among original positives, \context{} marks interpretive support that was over-counted as causal. Among original negatives, it marks interpretive support that was discarded as irrelevant. This two-sided movement rules out a simple missing-positive account and also rules out treating \context{} as a weak \cause{} label. The inherited binary boundary preserves many direct-cause decisions, but it cannot keep causal status and explanatory relevance separate.

\begin{table}[t]
  \centering
    \caption{Checks showing that \context{} is not a weak cause, annotation artifact, or distance heuristic.}
  \label{tab:boundary-checks-main}
  \scriptsize
  \setlength{\tabcolsep}{3pt}
  \resizebox{\linewidth}{!}{%
  \begin{tabular}{p{0.39\linewidth}p{0.57\linewidth}}
    \toprule
    Ruled-out explanation & Audit evidence \\
    \midrule
    \context{} is only missing positives or weak \cause{} &
    \context{} appears in both original positives (8.8\%) and original negatives (4.6\%). \\
    \addlinespace[2pt]
    \context{} is an annotation artifact &
    All 4,346 \context{} pairs have explicit non-default provenance. \\
    \addlinespace[2pt]
    \context{} is just a distance heuristic &
    68.4\% occur at distance $\geq3$, but 1,375 occur at shorter distances. \\
    \bottomrule
  \end{tabular}
  }
\end{table}

Table~\ref{tab:boundary-checks-main} rules out three simpler explanations for this boundary-crossing pattern. Together, these checks show that \context{} is neither a weak cause, nor a default annotation artifact, nor a distance-defined label. The boundary crossing pattern is therefore better explained as compression of explanatory relevance by the inherited binary pair/non-pair boundary. Full construction diagnostics, including source-binary preservation and self-pair behavior, are in Appendix~\ref{app:final-file-checks}.

\subsection{Boundary Uncertainty}
\label{sec:boundary-uncertainty}

\begin{figure}[t]
  \centering
  \includegraphics[width=0.95\linewidth]{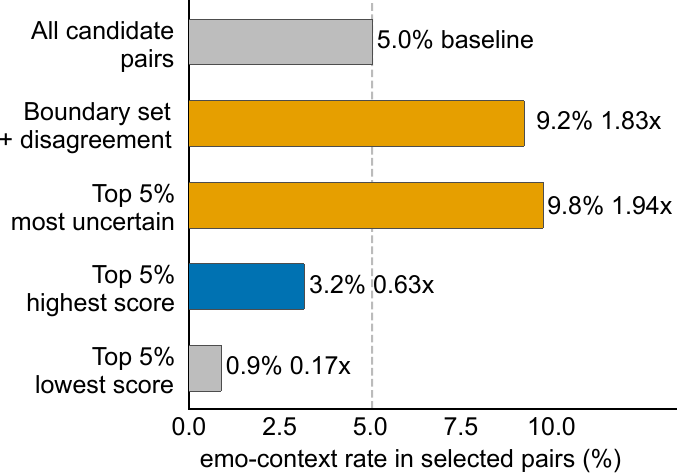}
  \caption{\context{} is enriched near binary uncertainty rather than at confident score extremes.}
  \label{fig:boundary-uncertainty}
\end{figure}

Figure~\ref{fig:boundary-uncertainty} tests P3 using a binary-source RWC-Fusion model with RoBERTa, WavLM, and CLIP features over all 86,075 scored pairs. \context{} appears most often where the binary model is least certain. Its overall rate is 5.0\%, but rises to 9.2\% in the boundary set with source disagreement and 9.8\% among the top 5\% most uncertain pairs. It is much rarer at the confident extremes, appearing in only 3.2\% of the top 5\% highest-score pairs and 0.9\% of the top 5\% lowest-score pairs.

This rules out two simpler explanations. \context{} is not just weak \cause{}, because it does not concentrate among high-score pairs. It is also not ordinary \nonpair{} noise, because it does not concentrate among low-score pairs. Instead, it appears where the binary model is unsure. These cases show where the binary label space absorbs discourse evidence without giving it a name.

\begin{table*}[!t]
  \centering
  \caption{Representative role-wise model behavior over three seeds. Ctx Hit is the \% of gold \context{} pairs predicted correctly, and Ctx$\rightarrow$NP is the \% predicted as \nonpair{}. Full model-modality results and cross-family collapse diagnostics are in Appendices~\ref{app:family-confusion} and ~\ref{app:full-results}.}
  \label{tab:diagnostic-main-results}
  \small
  \resizebox{\linewidth}{!}{%
  \begin{tabular}{llrrrrrrr}
    \toprule
    Model family & Mod. & Macro F1 & \cause{} F1 & \context{} F1 & \nonpair{} F1 & Ctx Hit & Ctx$\rightarrow$Cause & Ctx$\rightarrow$NP \\
    \midrule
    RoBERTa & T & 48.77$_{\pm0.39}$ & 40.28$_{\pm0.52}$ & 17.18$_{\pm1.47}$ & 88.84$_{\pm0.29}$ & 18.3 & 8.9 & 72.8 \\
    WavLM & A & 40.63$_{\pm0.33}$ & 28.26$_{\pm0.23}$ & 3.65$_{\pm0.73}$ & 89.97$_{\pm0.10}$ & 2.0 & 13.1 & 84.9 \\
    CLIP & V & 30.19$_{\pm1.32}$ & 18.50$_{\pm0.22}$ & 5.20$_{\pm2.02}$ & 66.86$_{\pm3.36}$ & 7.8 & 39.5 & 52.8 \\
    RWC-Fusion & T+A+V & 48.41$_{\pm0.50}$ & 40.89$_{\pm0.63}$ & 16.94$_{\pm1.74}$ & 87.41$_{\pm0.52}$ & 21.3 & 9.5 & 69.1 \\
    \mecpetwostep{} & T+A & 55.60$_{\pm0.08}$ & 52.98$_{\pm0.11}$ & 25.14$_{\pm0.91}$ & 88.67$_{\pm0.60}$ & 35.5 & 9.7 & 54.8 \\
    HiLo & T+A+V & 52.94$_{\pm0.61}$ & 50.60$_{\pm0.69}$ & 19.60$_{\pm2.06}$ & 88.61$_{\pm0.99}$ & 25.0 & 14.3 & 60.8 \\
    \mthreehg & T+A & 57.95$_{\pm0.07}$ & 56.73$_{\pm0.18}$ & 26.84$_{\pm0.33}$ & 90.30$_{\pm0.52}$ & 34.0 & 11.0 & 55.0 \\
    \bottomrule
  \end{tabular}
  }
  \vspace{-0.5em}
\end{table*}
\subsection{Source-Binary Compatibility}
\label{sec:source-binary-compatibility}

\begin{figure}[t]
  \centering
  \includegraphics[width=0.95\linewidth]{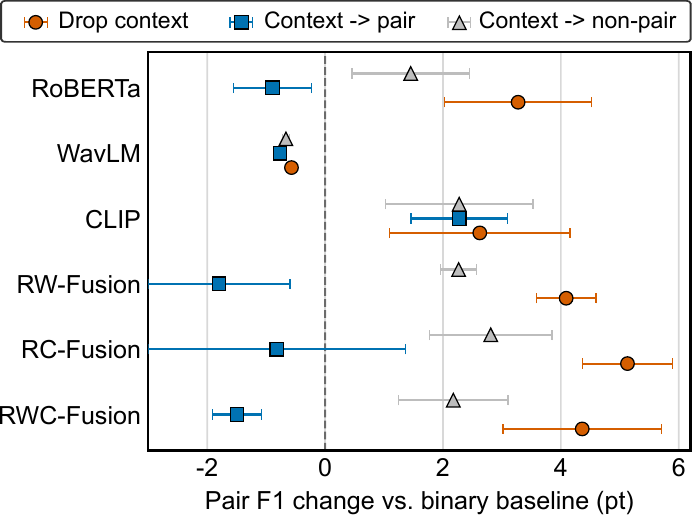}
  \caption{Pair F1 change after remapping three-role predictions to the binary ECPE task. Error bars show variation over three seeds.}
  \label{fig:source-binary-remap-main}
\end{figure}

Before using the role-aware labels to question explanation claims, we check whether they preserve the original ECPE task. Sections~\ref{sec:source-binary-transitions} and~\ref{sec:boundary-uncertainty} show that \context{} is compressed by the inherited binary boundary. Figure~\ref{fig:source-binary-remap-main} addresses a natural concern about whether three-role supervision gains interpretability by drifting from the source-binary objective. We remap three-role predictions to pair/non-pair labels and compare Pair F1 against models trained directly on original binary labels.

Role-aware supervision remains competitive with, and often improves over, direct binary training after remapping. The strongest results occur when the projection keeps the \cause{} channel separate, either by dropping predicted \context{} or by mapping it to \nonpair{}. Mapping predicted \context{} into the positive pair class often hurts text-grounded settings, showing that the gains do not come from expanding the positive class with contextual support, but from separating direct-cause evidence from non-triggering discourse evidence during training.

Thus, role-aware supervision does not reject binary ECPE. It isolates the part that measures direct-cause extraction from the part often over-read as broader explanation. The three-role label space remains compatible with source-binary evaluation while clarifying what binary pair scores can and cannot justify. Absolute source-binary remapping results are in Appendix~\ref{app:source-binary-remap}.

\subsection{Model Diagnostic Setup}
\label{sec:model-diagnostic-setup}

To test whether models preserve the audited role distinction, we evaluate text (T), audio (A), and video (V) baselines using RoBERTa, WavLM, and CLIP-derived features \citep{liu-etal-2019-roberta,chen2022wavlm,radford-etal-2021-learning}, along with stronger conversational and multimodal ECPE-style systems, including HiLo, \mecpetwostep{}, and \mthreehg{} \citep{wang2023multimodal,li2025multimodal,liang-etal-2025-m3hg}. Each setting uses seeds 42, 123, and 456; tables report mean and standard deviation. As the roles are imbalanced, evaluation emphasizes macro F1, class-wise F1, and role-specific diagnostics rather than accuracy. See Appendix~\ref{app:implementation-details} for full optimization, architecture recipes and compute details.

\section{Models Preserve Direct Causes More Reliably than Context}
\label{sec:model-preserve}

\subsection{Role-Wise Results}

Table~\ref{tab:diagnostic-main-results} gives the model-side counterpart of the role audit; calibration baselines are in Appendix~\ref{app:calibration-baselines}. Three-class pair-role prediction is learnable, and stronger ECPE-style systems improve aggregate macro F1. The more important pattern is role-wise. \cause{} and \nonpair{} are recovered much more reliably than \context{}, which remains the limiting role across model families.

The error direction matters. Among the representative rows, \mthreehg{} is the strongest overall setting, with the highest Macro F1 and the highest \context{} F1. Yet it predicts only 34.0\% of gold \context{} pairs correctly, while assigning 55.0\% to \nonpair{} and only 11.0\% to \cause{}. If \context{} were merely weak \cause{}, missed cases should move toward \cause{}. Instead, most missed contexts collapse into \nonpair{}, suggesting that models often treat contextual support as unsupported evidence. Appendix~\ref{app:family-confusion} shows the same \context{}-to-\nonpair{} tendency across model families.

\begin{figure}[t]
  \centering
  \includegraphics[width=0.9\linewidth]{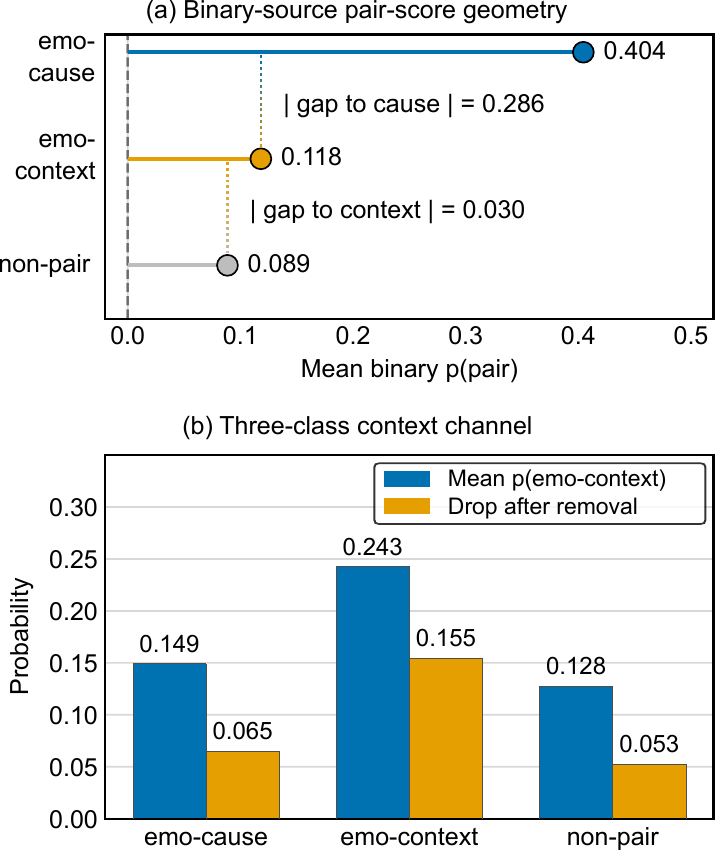}
  \caption{Context-channel diagnostics for RWC-Fusion. Binary supervision places gold \context{} near \nonpair{}, while three-class supervision exposes a weak context-sensitive signal.}
  \label{fig:context-channel-main}
\end{figure}

This pattern is not explained by modality choice alone. Although aggregate performance varies across text, audio, video, and fusion settings, the central failure mode is stable. Models preserve \cause{} and \nonpair{} distinctions more reliably than non-triggering discourse evidence. The issue is not simply adding more modality signal, but preserving \context{} as distinct from both direct causality and ordinary \nonpair{} evidence.

\subsection{Binary Scores Lack a Context Channel}
Figure~\ref{fig:context-channel-main} tests P4 by locating gold \context{} on the score axis. Under original binary supervision, \context{} is much closer to \nonpair{} than to \cause{} in mean $p(\mathrm{pair})$. This is expected under a pair/non-pair objective, since contextual support has no separate evidence channel and is mostly scored as absence of direct-cause evidence.

\begin{figure}[t]
  \centering
  \includegraphics[width=0.95\linewidth]{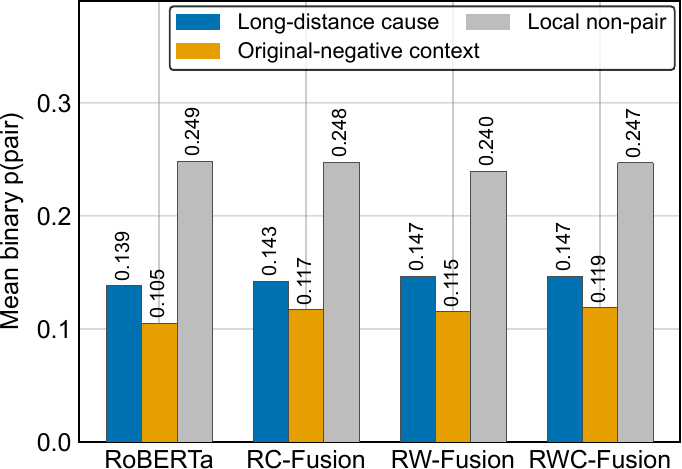}
  \caption{Binary pair scores favor shortcut-compatible \nonpair{} controls over evidence supported but structurally harder \cause{} and \context{} pairs.}
  \label{fig:shortcut-conflict-family}
\end{figure}

Three-class supervision changes the geometry, but only partially. Gold \context{} receives the highest mean \context{} probability, and candidate removal produces the largest \context{}-probability drop for gold \context{} pairs. This shows that the role is learnable when exposed by the label space. Yet the signal remains fragile, as the three-class RWC-Fusion model predicts \context{} for only 21.3\% of gold \context{} pairs, while 69.1\% are predicted as \nonpair{}. Thus, \context{} is not simply noise or impossible to learn; exposing a context channel does not ensure that models preserve it reliably. Appendix~\ref{app:binary-axis-removal-family} repeats these diagnostics across text-grounded and fusion controls.

\section{When Binary Pair Scores Favor Shortcut-Compatible Attributions}
\label{sec:shortcut-main}

\subsection{Natural Shortcut vs. Evidence Stress Test}

We instantiate P5 with a natural stress test that does not edit utterances. The shortcut-compatible control consists of final-labeled \nonpair{} candidates that are local to the target and lexically overlapping with it. The evidence-supported cases are long distance \cause{} pairs and original-negative \context{} pairs. This contrast reverses the usual alignment between evidence and easy structural cues. The unsupported control has locality and lexical overlap, while the evidence-supported cases are structurally harder.

Figure~\ref{fig:shortcut-conflict-family} shows the same ordering across binary-source controls. Local lexical \nonpair{} candidates receive the highest mean pair scores, above both long-distance \cause{} pairs and original-negative \context{} pairs. The latter two subsets contain final-role evidence for \cause{} or \context{}, but lack the shortcut compatible locality and overlap of the control. This does not show that the models ignore utterance evidence. It shows that binary pair scores can reward shortcut compatible plausibility over evidence supported but structurally harder pairs.

This is where the proxy stops behaving like explanation evidence. A high pair score can reflect a convenient attribution pattern rather than stronger evidence for why the emotion occurs. See Appendix~\ref{app:shortcut-vs-evidence} for full results.

\subsection{Matched Context Non-Pair Controls}
\label{sec:matched-controls}

The shortcut vs. evidence stress test shows that unsupported local candidates can score highly when they align with locality and lexical overlap. A stricter test asks whether contextual support remains separable after these cues are controlled. For each gold \context{} pair, we select a final-labeled \nonpair{} control matched on distance, source-binary metadata, speaker relation, target emotion, candidate length, lexical overlap, and dialogue type. We then compare matched-pair score gaps under binary-source and three-class controls. This design tests whether models preserve a context signal once obvious structural and lexical explanations no longer distinguish the roles.

\begin{figure}[t]
  \centering
  \includegraphics[width=0.9\linewidth]{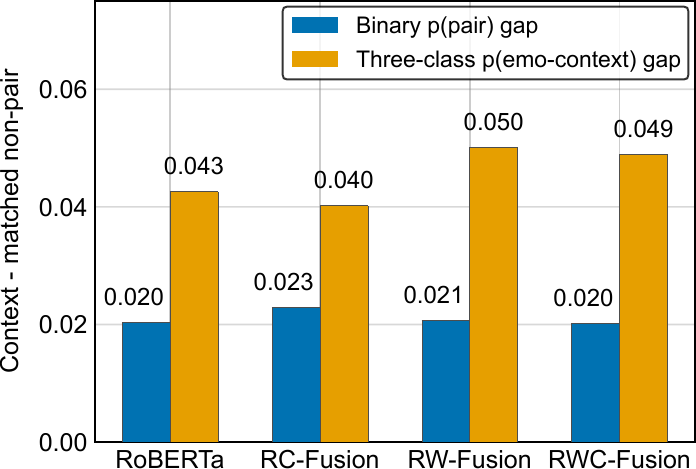}
  \caption{Matched \context{}-\nonpair{} score gaps. Positive gaps indicate higher scores for gold \context{} pairs than matched \nonpair{} controls.}
  \label{fig:matched-context-controls}
\end{figure}

Figure~\ref{fig:matched-context-controls} shows a consistent but narrow separation. Gold \context{} pairs receive slightly higher scores than their matched \nonpair{} controls, with binary $p(\mathrm{pair})$ gaps around 0.020 - 0.023 and three-class $p(\context{})$ gaps around 0.040 - 0.050. The positive gaps show that \context{} is not simply noise or an unlearnable residual class. Their small magnitude is the diagnostic point. Once shortcut-compatible cues are matched away, the evaluated models retain only a weak signal for non-triggering discourse evidence. The validity risk is not only that unsupported local candidates can be over-ranked, but also that the boundary between contextual support and unsupported plausibility remains weakly represented.

\subsection{Observable Shortcut Probes}
The stress test provides one controlled contrast. We next ask whether observable structure predicts binary-source behavior more broadly. For each trained model, we fit lightweight cross-validated probes using distance, turn position, speaker relation, target and candidate emotions, dialogue metadata, length buckets, and lexical overlap. These probes exclude utterance text and multimodal content, measuring how much model behavior is recoverable from candidate-pair structure alone.

Table~\ref{tab:shortcut-proxy-main} shows that these observable features predict binary-source hard pair decisions with AUC between 0.833 and 0.855, and recover pair-score rankings with Spearman correlations between 0.462 and 0.605. This does not imply that ECPE models ignore evidence. It shows that a substantial part of binary-source behavior is reachable from structural and lexical cues alone. The validity risk is therefore not limited to one stress subset. Binary ECPE evaluation can reward structurally plausible pair decisions without requiring a stable channel for non-triggering discourse evidence.

\paragraph{Distance is a risk factor, not the role definition.}
Distance explains why locality matters, but it does not define \context{}. Appendix~\ref{app:distance-distribution} shows that \cause{} concentrates at short distances, while \context{} shifts longer. Final file diagnostics show that 68.4\% of \context{} pairs occur at distance $\geq3$, but 1,375 occur at shorter distances, and same distance candidates can differ in role. We therefore treat distance as a shortcut risk factor rather than an annotation rule.

Together with the stress test, this shows why binary pair scores should not be read as explanatory evidence by default. They partly track structural plausibility, while the channel for non-triggering discourse evidence remains weak. See Appendix~\ref{app:proxy-only} for full proxy-only results, including source-binary and three-class targets.

\begin{table}[t]
  \centering
  \caption{Observable shortcut probes for binary-source behavior. AUC/Acc. evaluate hard pair decisions, and $R^2$/$\rho$ evaluate pair-score prediction.}
  \label{tab:shortcut-proxy-main}
  \scriptsize
  \setlength{\tabcolsep}{2pt}
  \resizebox{\linewidth}{!}{%
  \begin{tabular}{lrrrr}
    \toprule
    Model & AUC & Acc. & $R^2$ & $\rho$ \\
    \midrule
    RoBERTa & 0.855$_{\pm0.007}$ & 79.6$_{\pm1.1}$ & 0.282$_{\pm0.058}$ & 0.548$_{\pm0.063}$ \\
    RW-Fusion & 0.851$_{\pm0.025}$ & 80.1$_{\pm2.7}$ & 0.295$_{\pm0.058}$ & 0.605$_{\pm0.054}$ \\
    RC-Fusion & 0.850$_{\pm0.020}$ & 79.1$_{\pm2.0}$ & 0.269$_{\pm0.030}$ & 0.462$_{\pm0.114}$ \\
    RWC-Fusion & 0.833$_{\pm0.028}$ & 78.8$_{\pm2.3}$ & 0.263$_{\pm0.079}$ & 0.562$_{\pm0.085}$ \\
    \bottomrule
  \end{tabular}}
\end{table}

\section{Discussion}
\label{sec:discussion}

\paragraph{Scope of the claim.}
The results do not invalidate binary ECPE. When the task is direct-cause extraction, a pair/non-pair output remains meaningful if described as such. The problem begins when the same score is used as evidence that a model explains why the target emotion occurred. In that setting, a system may recover many direct causes while collapsing \context{} into \nonpair{} and ranking shortcut-compatible unsupported candidates above harder evidence supported pairs. The lesson is not that binary ECPE is broken; it is that binary pair scores have been asked to certify more than they measure.

\paragraph{Applied relevance.}
This distinction matters because deployed dialogue systems often turn pair scores into user-facing explanations. In customer-service analysis, a returned package may be the direct trigger, while prior complaints or unresolved service history explain why the frustration is severe. In healthcare, HR, education, and legal dialogue analysis, earlier disclosures, authority relations, prior feedback, or interaction history may shape emotion interpretation without being the immediate cause. In such settings, binary pair scores should be treated as direct-cause signals, not as evidence grounded emotion explanations.

\paragraph{How \dataset{} should be used.}
\dataset{} should be used as a diagnostic instrument for a label-space bottleneck. It is not a universal ontology of emotional explanation, and it does not assume that every emotion has an observable dialogue cause. The \context{} criterion is intentionally strict, such that it requires observable support whose removal weakens interpretation of the target emotion. This keeps \context{} from becoming a catch-all label for preceding utterances.

\paragraph{Evaluation recommendation.}
Future ECPE evaluations should report direct-cause extraction and explanation diagnostics separately. Pair F1 is appropriate for the former, but it should not be the only evidence for claims about causal reasoning in dialogue. Explanation oriented systems should add role-wise metrics, boundary uncertainty analysis, shortcut vs. evidence stress tests, and error analyses that separate unsupported local plausibility from discourse evidence.

\section{Related Work}
\label{sec:related}

\paragraph{Emotion-cause extraction.}
ECPE was introduced as a pair extraction task over emotions and causes \citep{xia-ding-2019-emotion}. Subsequent ECPE work improved pair modeling through two-dimensional representations and inter-clause interaction \citep{ding-etal-2020-ecpe,wei-etal-2020-effective}, as well as alternative training or decoding formulations such as sliding-window learning, sequence labeling, and boundary-aware constraints \citep{ding-etal-2020-end,yuan-etal-2020-emotion,feng-etal-2023-joint}. More recent work has explored prompting and knowledge use, fine-grained clue learning, conversational event reasoning, and unified emotion-cause analysis across related tasks \citep{gu-etal-2024-emoprompt,yu-etal-2024-mgcl,wang-etal-2024-enhancing-emotion,yu-etal-2025-one}. These approaches improve extraction within ECPE-style pair/non-pair label spaces. Our work asks what explanation claims such label spaces can support.

\paragraph{Conversational and causal emotion analysis.}
Conversational emotion-cause analysis extends ECPE from documents to dialogue, speaker interaction, and conversational evidence \citep{poria2021recognizing,li2023ecpec,jeong-bak-2023-conversational}. Causal emotion entailment further highlights the role of commonsense and discourse evidence in emotion reasoning \citep{zhao-etal-2023-knowledge}. We share this emphasis on dialogue evidence, but study a different question: \emph{whether a binary pair score can distinguish direct causality from contextual support}.

\paragraph{Multimodal emotion-cause and emotion recognition.}
Multimodal ECPE and multimodal emotion-cause analysis incorporate text, speech, and visual evidence \citep{wang2023multimodal,wang-etal-2024-semeval,li2025multimodal,liang-etal-2025-m3hg}. Broader emotion-recognition work in conversations approaches context and modality from related but different angles, including multimodal fusion and semantic refinement \citep{shi-huang-2023-multiemo,zhang-li-2023-cross}, temporal and cross-modal interaction modeling \citep{nguyen-etal-2023-conversation,yun-etal-2024-telme}, and recent work on graph-based structure, explicit evidence and clue modeling, causal prompting, and cross-space interaction \citep{ai-etal-2025-revisiting,zhang-tan-2025-ecerc,shen-etal-2025-coe,jing-etal-2026-causal,lyu-etal-2026-cross}. These works motivate the importance of context and modality. Our findings show that stronger contextual and multimodal modeling alone does not resolve a pair-label validity problem.

\paragraph{Context, bias, and diagnostic evaluation.}
Emotion-cause analysis has long considered context awareness and position bias \citep{li-etal-2018-co,yan-etal-2021-position}. More broadly, emotion modeling work shows that emotion-label granularity and annotation reliability affect what benchmark scores mean \citep{wen-etal-2023-learning,troiano-etal-2023-dimensional}. \dataset{} follows this diagnostic tradition, but focuses on an ECPE-specific bottleneck: \emph{whether the inherited binary label space compresses contextual evidence needed for explanation}.

\section{Conclusion}

Binary ECPE remains useful for direct-cause extraction, but its pair scores should not be treated as sufficient evidence for evidence grounded emotion explanation. Using \dataset{} as a diagnostic instrument, we show that \context{} crosses both sides of the inherited binary boundary, concentrates near binary uncertainty, and is often collapsed into \nonpair{}. A shortcut vs. evidence stress test further shows that unsupported local lexical \nonpair{} controls can outrank evidence supported long distance \cause{} and original-negative \context{} pairs. The lesson is not that binary ECPE is wrong; it is that binary pair scores answer a narrower question than explanation oriented systems often ask. Future evaluations should therefore separate direct-cause extraction from evidence grounded dialogue emotion explanation.

\section*{Limitations}

\dataset{} is designed as a diagnostic audit of an inherited ECPE label space. Its purpose is to test whether the pair/non-pair boundary compresses a separable discourse-evidence role, rather than to serve as a universal ontology of emotion explanation. The construction and validation combine LLM-assisted first-pass labeling, dialogue-level review, targeted human refinement, additional human review of high-risk boundary transitions, a blind human reannotation audit over boundary-focused packets, and final file consistency checks. These checks provide targeted evidence for the boundary-compression analysis.

The current audit is based on IEMOCAP-style dyadic dialogues, so future work can extend the same diagnostic protocol to spontaneous, multi-party, domain-specific, and culturally diverse interactions. The blind human reannotation audit supports reproducibility in difficult boundary regions, while larger-scale independent annotation may further clarify cases where background information can be interpreted as part of a causal chain. Finally, future systems may benefit from richer pragmatic, discourse, and interactional cues to better preserve contextual support.

\balance
\bibliography{references}

\clearpage
\appendix

\section{Diagnostic Instrument and Annotation Criteria}
\label{app:diagnostic-instrument}

This section documents the construction controls for \dataset{} while keeping the focus on the evaluation-validity claim of the paper. The audit is not intended to replace existing ECPE benchmarks. It fixes the inherited candidate space, preserves source-binary metadata, and adds pair-role labels so that boundary compression can be inspected directly. 

\subsection{Diagnostic Workflow}
\label{app:diagnostic-workflow}

\begin{figure}[ht]
  \centering
  \includegraphics[width=\linewidth]{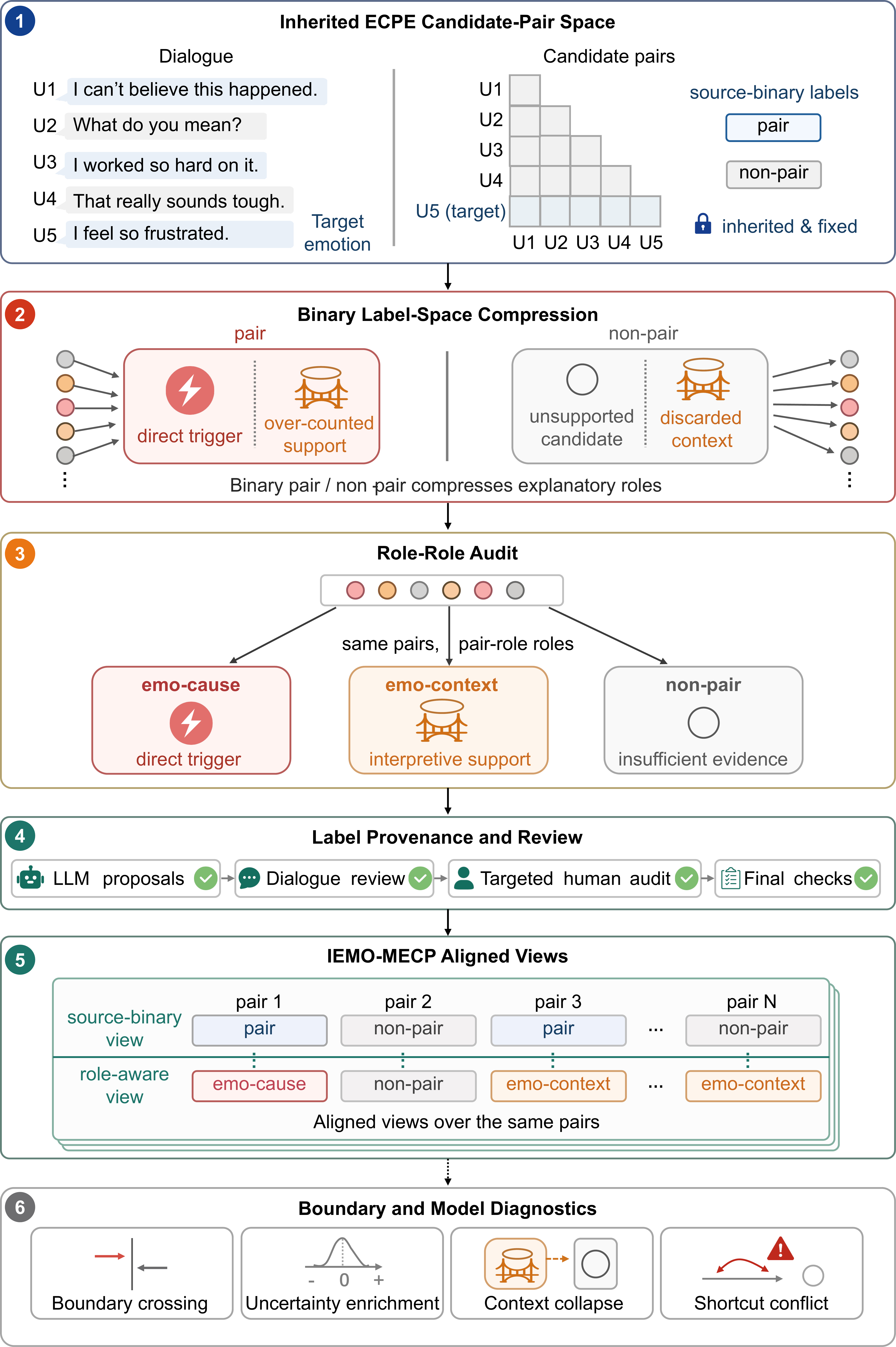}
  \caption{Construction-side diagnostic workflow for \dataset{}. The candidate space stays fixed, while role labels and source-binary metadata make boundary compression measurable.}
  \label{fig:audit-workflow}
\end{figure}

Figure~\ref{fig:audit-workflow} summarizes the construction-side logic. The workflow starts from the label-space bottleneck where binary ECPE scores place direct causes, contextual support, and unsupported candidates on one pair/non-pair axis. The boundary-diagnosis stage then targets regions where that axis is likely to hide non-triggering discourse evidence. Pair-role auditing separates the same candidate pairs into \cause{}, \context{}, and \nonpair{} while retaining the original source-binary label for each pair. The resulting data allow binary and role-aware views to be compared over the same lower-triangular dialogue-pair space.

The workflow is diagnostic rather than promotional. Its purpose is to make boundary compression measurable, such that if \context{} crosses the inherited boundary, concentrates near binary uncertainty, or collapses back into \nonpair{} under model scoring, these patterns can be interpreted as evidence about what binary pair scores do and do not measure.

\subsection{Label Provenance and Targeted Human Judging}
\label{app:human-validation}

The final labels are not unchecked LLM outputs. The LLM-assisted stage supplies structured role proposals and rationales, but the released labels retain explicit provenance for how each candidate enters the final files. Table~\ref{tab:label-provenance} gives the final-file provenance distribution. The key diagnostic point is that all final \context{} pairs have non-default provenance, and all are associated with targeted human judging. Thus, \context{} is not introduced by the default lower-triangular \nonpair{} fill, by a distance rule, or by an unreviewed automatic pass.

\begin{table*}[t]
  \centering
  \caption{Final pair-role label provenance. Counts are final released labels.}
  \label{tab:label-provenance}
  \small
  \setlength{\tabcolsep}{5pt}
  \begin{tabular}{lrrrr}
    \toprule
    Provenance source & Total & \cause{} & \context{} & \nonpair{} \\
    \midrule
    Default lower-triangular fill & 70,649 & 0 & 0 & 70,649 \\
    LLM-assisted retained labels & 3,631 & 3,631 & 0 & 0 \\
    Targeted human judge & 11,427 & 5,004 & 4,346 & 2,077 \\
    Human refinement actions & 368 & 0 & 0 & 368 \\
    \midrule
    Final total & 86,075 & 8,635 & 4,346 & 73,094 \\
    \bottomrule
  \end{tabular}
\end{table*}

\begin{table*}[t]
  \centering
  \caption{Targeted human-judge inventory. Reviewed regions are boundary cases where source labels, LLM proposals, and consistency checks are likely to diverge.}
  \label{tab:human-judge-inventory}
  \small
  \setlength{\tabcolsep}{3.5pt}
  \resizebox{\linewidth}{!}{%
  \begin{tabular}{lrrrrrrr}
    \toprule
    Human-judge region & Reviewed & \multicolumn{3}{c}{Manual decision in review file} & \multicolumn{3}{c}{Final labels with this provenance} \\
    \cmidrule(lr){3-5}\cmidrule(lr){6-8}
     & pairs & \cause{} & \context{} & \nonpair{} & \cause{} & \context{} & \nonpair{} \\
    \midrule
    Multi-cause consistency review & 2,678 & 2,428 & 250 & 0 & 2,428 & 201 & 0 \\
    Single-cause consistency review & 2,082 & 2,082 & 0 & 0 & 2,001 & 0 & 0 \\
    Nearby-context consistency review & 448 & 0 & 448 & 0 & 0 & 448 & 0 \\
    Original negative proposed as \cause{} & 2,517 & 327 & 118 & 2,072 & 213 & 118 & 2,047 \\
    Original negative proposed as \context{} & 2,975 & 0 & 2,969 & 6 & 0 & 2,969 & 6 \\
    Original positive proposed as \context{} & 724 & 292 & 432 & 0 & 78 & 432 & 0 \\
    Original positive proposed as \nonpair{} & 2,138 & 2,016 & 79 & 43 & 96 & 60 & 24 \\
    Source-positive recovery after lost cause coverage & 306 & 188 & 118 & 0 & 188 & 118 & 0 \\
    \bottomrule
  \end{tabular}
  }
\end{table*}

\begin{table*}[!t]
  \centering
  \caption{Boundary-focused packets used in the blind human reannotation audit. Packet settings and role contrasts are used only for sampling and are not shown to annotators.}
  \label{tab:blind-human-subset}
  \small
  \setlength{\tabcolsep}{2pt}
  \resizebox{\linewidth}{!}{%
  \begin{tabular}{llrlr}
    \toprule
    Packet setting & Role contrast & Packets & Candidate composition & Judgments \\
    \midrule
    Two-role & \cause{} / \context{} & 40 & 40 \cause{} + 40 \context{} & 80 \\
    Two-role & \context{} / \nonpair{} & 40 & 40 \context{} + 40 \nonpair{} & 80 \\
    Two-role & \cause{} / \nonpair{} & 40 & 40 \cause{} + 40 \nonpair{} & 80 \\
    Three-role & \cause{} / \context{} / \nonpair{} & 30 & 30 \cause{} + 30 \context{} + 30 \nonpair{} & 90 \\
    \midrule
    Total & All contrasts & 150 & 110 \cause{} + 110 \context{} + 110 \nonpair{} & 330 \\
    \bottomrule
  \end{tabular}
  }
\end{table*}

Human judging is targeted rather than uniform reannotation of every candidate pair. This design matches the validity question. Review is concentrated where the main claims depend on boundary distinctions. Original positives proposed away from direct causality, original negatives proposed as explanatory, targets whose inherited cause coverage becomes inconsistent after refinement, and candidate regions where \context{} could otherwise be confused with proximity or weak causality. The ``reviewed pairs'' column in Table~\ref{tab:human-judge-inventory} counts rows in the review files. The final-provenance columns count released labels whose final source points to that review region. These counts need not match exactly because later consistency decisions and overlapping review regions can supersede earlier rows.

\subsection{Blind Human Re-annotation Audit}
\label{app:blind-human-reannotation}

\paragraph{Audit design.}
The targeted human-judge inventory in Appendix~\ref{app:human-validation} documents where final labels receive explicit human provenance. We additionally test whether the central pair-role boundary can be recovered under independent blind judgment. The audit uses target-centered packets where each packet contains one target emotion utterance and two or three candidate utterances. Annotation remains pair-level, with one candidate-to-target judgment per row.

The audit is boundary-focused rather than distribution-matching. We sample packets without reusing the same target utterance, prioritize high-overlap or distance-matched boundary cases, and include candidates from targeted human-judge regions when available. The hidden final labels are balanced across \cause{}, \context{}, and \nonpair{}, but these labels are used only for audit construction. Table~\ref{tab:blind-human-subset} summarizes the packet design.

\paragraph{Blind annotation setting.}
Three annotators independently label all candidate-to-target judgments. They see the dialogue context, the target emotion utterance, the target emotion label, and the candidate utterance. They do not see the final pair-role label, the source-binary label, the label provenance, the LLM proposal or rationale, or the packet type. Agreement therefore measures independent recovery of the role boundary rather than agreement with a visible proposal.

\begin{table*}[!t]
  \centering
  \caption{Agreement in the blind human reannotation audit. Raw agreement is averaged over annotator pairs. Class-conditioned columns use the hidden final label only for analysis. These labels are not visible during annotation.}
  \label{tab:blind-human-agreement}
  \small
  \begin{tabular}{lrrrr}
    \toprule
    Metric & Overall & \cause{} & \context{} & \nonpair{} \\
    \midrule
    Raw pairwise agreement & 0.821 & 0.811 & 0.767 & 0.893 \\
    Mean pairwise Cohen's $\kappa$ & 0.732 & -- & -- & -- \\
    Pairwise Cohen's $\kappa$ range & 0.696--0.758 & -- & -- & -- \\
    Krippendorff's $\alpha$ & 0.732 & -- & -- & -- \\
    3/3 unanimous judgments & 245 / 330 & 83 / 110 & 75 / 110 & 87 / 110 \\
    2/3 majority judgments & 78 / 330 & 33 / 110 & 32 / 110 & 13 / 110 \\
    \bottomrule
  \end{tabular}
\end{table*}

\paragraph{Agreement pattern.}
Table~\ref{tab:blind-human-agreement} shows that the role boundary remains reproducible even on packets selected to stress difficult distinctions. The lowest agreement appears for \context{}, as expected for the boundary role, but it still receives stable majority behavior under blind annotation. The result supports the use of \context{} as a recoverable discourse-evidence role rather than as an unchecked artifact of the construction pipeline.

\begin{table}[t]
  \centering
  \caption{Disagreement patterns in the blind human reannotation audit. Counts are computed over non-unanimous judgments.}
  \label{tab:blind-human-disagreement}
  \small
  \setlength{\tabcolsep}{5pt}
  \begin{tabular}{lrr}
    \toprule
    Disagreement type & Count & \% \\
    \midrule
    \cause{} vs. \context{} & 52 & 61.2 \\
    \context{} vs. \nonpair{} & 19 & 22.4 \\
    \cause{} vs. \nonpair{} & 7 & 8.2 \\
    All three labels different & 7 & 8.2 \\
    \midrule
    Total & 85 & 100.0 \\
    \bottomrule
  \end{tabular}
\end{table}

\paragraph{Diagnostic interpretation.}
Table~\ref{tab:blind-human-disagreement} shows that the disagreement structure is consistent with the role hierarchy in Section~\ref{sec:framework}. Most ambiguity is between direct-cause evidence and non-triggering discourse evidence, not between \context{} and ordinary \nonpair{} evidence. The audit therefore does not claim that every boundary case is trivial. It shows that the difficult boundary is structured, where annotators mostly disagree over whether discourse evidence is directly causal or contextual, rather than over whether \context{} is merely unsupported proximity.

\subsection{LLM-Assisted Draft-to-Final Changes}
\label{app:llm-human-change}

\begin{table*}[!t]
  \centering
  \small
  \setlength{\tabcolsep}{4pt}
  \caption{OpenAI Codex draft labels compared with final labels after targeted human audit. Draft \nonpair{} includes pairs not elevated by the LLM-assisted stages.}
  \label{tab:gpt-human-change}
  \resizebox{\linewidth}{!}{%
  \begin{tabular}{lrrrrrrrr}
    \toprule
    Scope & Pairs & \multicolumn{3}{c}{OpenAI Codex draft} & \multicolumn{3}{c}{Final after human audit} & Changed \\
    \cmidrule(lr){3-5}\cmidrule(lr){6-8}
     & & \cause{} & \context{} & \nonpair{} & \cause{} & \context{} & \nonpair{} & \\
    \midrule
    Overall & 86,075 & 8,579 & 3,699 & 73,797 & 8,635 & 4,346 & 73,094 & 5,313 (6.17\%) \\
    Original positives & 9,145 & 6,062 & 724 & 2,359 & 8,316 & 805 & 24 & 2,688 (29.39\%) \\
    Original negatives & 76,930 & 2,517 & 2,975 & 71,438 & 319 & 3,541 & 73,070 & 2,625 (3.41\%) \\
    \bottomrule
  \end{tabular}
  }
\end{table*}

Table~\ref{tab:gpt-human-change} compares the OpenAI Codex draft layer, produced with the \texttt{gpt-5.5 xhigh} model/reasoning setting, with the final labels after targeted human audit. The draft is defined over the full candidate space, where pairs not elevated by the LLM-assisted stages are counted as draft \nonpair{}. The ``Changed'' column therefore measures draft-to-final movement after audit, not an LLM error rate.

The movement pattern is diagnostic. Original positives change more often because the audit restores direct-cause coverage and separates over-broad positives from non-triggering support. Original negatives remain comparatively stable, while a targeted subset is elevated to \context{} or \cause{} when the dialogue supplies evidence. This supports the intended use of \dataset{}, where the audit refines high-risk boundary regions while preserving the inherited binary task structure.

\subsection{Final-File Diagnostic Checks}
\label{app:final-file-checks}

\begin{table*}[t]
  \centering
  \caption{Extended final-file diagnostics supporting the boundary-compression analysis.}
  \label{tab:construction-diagnostics}
  \small
  \resizebox{\linewidth}{!}{%
  \begin{tabular}{p{0.22\linewidth}p{0.35\linewidth}p{0.35\linewidth}}
    \toprule
    Diagnostic & Final-file evidence & Interpretation \\
    \midrule
    Binary task is preserved & 8,316/9,145 original positives remain \cause{} (90.9\%); 73,070/76,930 original negatives remain \nonpair{} (95.0\%). & The role-aware view refines the source ECPE boundary rather than replacing it with an unrelated task. \\
    \addlinespace[2pt]
    Context crosses the inherited boundary & 805/9,145 original positives (8.8\%) and 3,541/76,930 original negatives (4.6\%) become \context{}. & \context{} is not only a missing-positive phenomenon and not only a weakened cause label. \\
    Context labels have explicit provenance & 4,346/4,346 final \context{} pairs have explicit non-default provenance in the released label files. & The context class is not produced by the default \nonpair{} fill or by a distance-only rule. \\
    \addlinespace[2pt]
    Self-pairs are not automatic causes & Self-pairs split into 2,794 \cause{} and 4,639 \nonpair{} cases, with 0 \context{} cases. & A self-pair is accepted as \cause{} only when the target utterance itself contains sufficient causal or appraisal evidence. \\
    \addlinespace[2pt]
    Distance is a risk factor, not a rule & 2,971/4,346 \context{} pairs occur at distance $\geq 3$ (68.4\%), but 1,375 occur at shorter distances. & \context{} is distance-skewed, yet still requires semantic evidence specific to the target emotion. \\
    \bottomrule
  \end{tabular}
  }
\end{table*}

\begin{table*}[t]
  \centering
  \caption{Representative transitions from the original binary space to the final pair role-aware view.}
  \label{tab:boundary-examples}
  \small
  \resizebox{\linewidth}{!}{%
  \begin{tabular}{p{0.10\linewidth}p{0.28\linewidth}p{0.24\linewidth}p{0.31\linewidth}}
    \toprule
    Transition & Target utterance & Candidate utterance & Role rationale \\
    \midrule
    Positive\newline$\rightarrow$ context &
    The campus is beautiful. I'm so excited. &
    It's safe around there. &
    The candidate adds situational background, but the target already expresses the emotion. \\
    \addlinespace[2pt]
    Positive\newline$\rightarrow$ context &
    I feel really bad for his family, you know. &
    Yeah, he kind of did a lot of things, but we acted together, sometimes. &
    The candidate adds background about shared responsibility, but the target already expresses the emotion. \\
    \addlinespace[2pt]
    Negative\newline$\rightarrow$ context &
    This is ridiculous. I seriously do not understand why you think these automated systems are supposed to work for anybody. &
    All right. I'm sorry, sir that you've been having a hard time. &
    The candidate supports interpretation of the target frustration, but it does not itself trigger the emotion. \\
    \addlinespace[2pt]
    Negative\newline$\rightarrow$ context &
    I didn't know him, Tricia, but uh- I heard that he was a great man. &
    Turned around for the worst, you know. &
    The candidate supplies interpretive background for the target's sadness, but it does not directly cause the emotion. \\
    \bottomrule
  \end{tabular}
  }
\end{table*}

Table~\ref{tab:construction-diagnostics} provides the main file-level sanity checks behind the diagnostic audit. The first row addresses task preservation, such that most source-positive pairs remain \cause{}, and most source-negative pairs remain \nonpair{}. This supports the claim that \dataset{} refines binary ECPE rather than replacing it with an unrelated labeling scheme.

The second row is the central boundary-compression evidence. \context{} appears among both source positives and source negatives. This two-sided pattern matters because it argues against two simpler interpretations, where \context{} is neither a set of missing positives, nor a weaker version of \cause{}. Instead, the same discourse role is forced onto both sides of the inherited binary boundary.

The remaining rows guard against common failure modes in interpreting the labels. Explicit provenance for all \context{} pairs shows that the class was not produced by default negative filling. The self-pair row shows that same-turn candidates are not automatically treated as direct causes; self-pairs require observable causal or appraisal evidence. The distance row shows that \context{} is distance-skewed but not distance-defined. Together, these checks support the main-paper claim that \context{} is a discourse-evidence role rather than a construction artifact.

\subsection{Boundary Examples}
\label{app:boundary-examples-section}

Table~\ref{tab:boundary-examples} illustrates the same discourse role moving through both sides of the inherited boundary. Positive-to-context cases show over-causalized support. The candidate helps interpret the target emotion, but the target utterance already contains the main emotional expression or appraisal. At the same time, negative-to-context cases show discarded discourse evidence where the inherited source label is negative, yet the candidate still makes the target emotion easier to interpret.

These examples make the linguistic role of \context{} explicit. The positive-to-context cases should not be counted as direct triggers merely because they are helpful. The negative-to-context cases should not be discarded merely because they are not triggers. This is the same validity problem studied in the main text, where \textbf{\emph{binary ECPE can keep many direct causes while still compressing evidence needed for grounded explanation}}.

\section{Formalizing Binary Boundary Compression}
\label{app:formal-compression}

The main text treats binary ECPE as a projection from a role-aware evidential space to a two-label decision space. Here we state the implication explicitly. Let \(\mathcal{R}=\{\cause{},\context{},\nonpair{}\}\), where \(\cause{}\) is direct-cause evidence, \(\context{}\) is non-triggering explanatory evidence, and \(\nonpair{}\) is insufficient evidence. Binary ECPE observes only \(y^{\mathrm{bin}}\in\{0,1\}\), so any binary evaluation induces a projection \(\pi:\mathcal{R}\rightarrow\{0,1\}\).

For direct-cause extraction, \(\pi(\cause{})=1\) and \(\pi(\nonpair{})=0\). The remaining role, \(\context{}\), is non-causal but explanatorily relevant. If \(\pi(\context{})=1\), the projection preserves explanatory relevance but loses causal status by evaluating contextual support as a cause. If \(\pi(\context{})=0\), the projection preserves non-causality but loses explanatory relevance by evaluating contextual support as irrelevant. Therefore, no binary projection that preserves the usual direct-cause interpretation can also preserve the distinction among direct causality, non-triggering explanatory support, and insufficient evidence.

This does not imply that binary ECPE is invalid. It implies that binary ECPE is valid for a narrower measurement target, which is direct-cause extraction. When the same binary score is interpreted as evidence of grounded explanation, the projection becomes under-specified. The empirical diagnostics in the main paper test whether this formal compression appears in the inherited ECPE candidate space and in model behavior.

\section{Training and Implementation Details}
\label{app:implementation-details}

This section fixes the implementation setting for the role-level diagnostics. The goal is not leaderboard optimization. It is to make clear which splits, model recipes, environments, and construction-support tools underlie the reported comparisons.

\subsection{Final \dataset{} Split Statistics}
\label{app:split-statistics}

Table~\ref{tab:appendix-split-stats} reports the released split structure and final pair-role counts. Raw dialogues are unique \texttt{raw\_doc\_id} units; chunks are the chunked dialogue instances in the split files; utterances are counted within those chunks. The table also confirms that the benchmark keeps the original lower-triangular candidate-pair space while evaluating three-class supervision over \cause{}, \context{}, and \nonpair{}.

\begin{table}[h]
  \centering
  \caption{Released split statistics for \dataset{}.}
  \label{tab:appendix-split-stats}
  \small
  \setlength{\tabcolsep}{4pt}
  \resizebox{\linewidth}{!}{%
  \begin{tabular}{lrrrr}
    \toprule
    Statistic & Train & Valid & Test & Overall \\
    \midrule
    Pairs & 53,323 & 14,488 & 18,264 & 86,075 \\
    \cause{} & 5,499 & 1,342 & 1,794 & 8,635 \\
    \context{} & 2,630 & 719 & 997 & 4,346 \\
    \nonpair{} & 45,194 & 12,427 & 15,473 & 73,094 \\
    \midrule
    Avg. chunks/dialog & 2.71 & 2.83 & 3.06 & 2.80 \\
    Avg. utterances/chunk & 17.64 & 17.99 & 17.08 & 17.57 \\
    Avg. pairs/chunk & 205.09 & 213.06 & 192.25 & 203.49 \\
    \bottomrule
  \end{tabular}
  }
\end{table}

The split statistics explain the evaluation choices in the main paper. The overall class distribution is strongly skewed toward \nonpair{}, which motivates macro F1 and class-wise F1 in the main evaluation. At the same time, \context{} contains 4,346 candidate pairs overall, so it is not merely an anecdotal boundary category. The same three-role structure is present in train, validation, and test splits, allowing \context{} behavior to be evaluated under the same candidate-space assumptions as \cause{} and \nonpair{}.

\subsection{\dataset{} Distance Distributions}
\label{app:distance-distribution}

\begin{figure*}[t]
  \centering
  \includegraphics[width=0.9\textwidth]{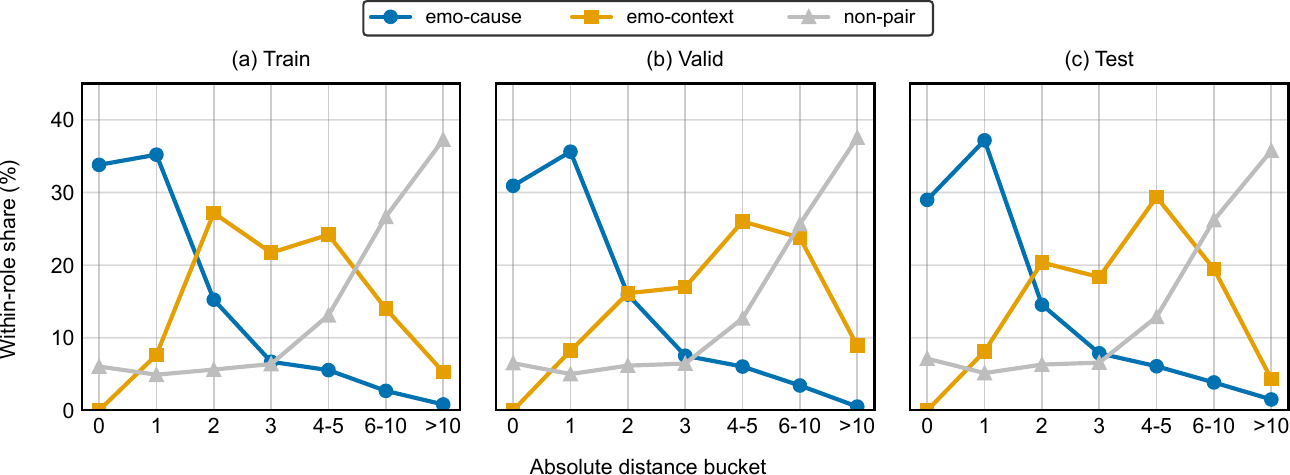}
  \caption{Role-conditioned absolute-distance distributions by split. \context{} shifts toward longer distances than \cause{}, but distance buckets overlap across roles.}
  \label{fig:appendix-distance-distribution}
\end{figure*}

Figure~\ref{fig:appendix-distance-distribution} shows that \context{} is shifted toward longer distances than \cause{} across splits. This helps explain why contextual support can collapse toward \nonpair{} in models biased toward local causes.

However, distance is not the role definition. The role distributions overlap, and shorter-distance \context{} pairs remain. Distance is therefore a shortcut risk factor, not an annotation rule. The semantic criterion is unchanged, such that \context{} must provide non-triggering support whose removal weakens interpretation of the target emotion.

\subsection{Optimization Settings}
\label{app:hparam-analysis}

All runs optimize the three-class pair-role task with class-weighted cross-entropy. Class weights are computed from the training split for the active task. RoBERTa, WavLM, CLIP, RW-Fusion, RC-Fusion, and RWC-Fusion share a single baseline recipe as follows. AdamW with learning rate $2\times10^{-5}$, training batch size 16, evaluation batch size 32, 10 epochs, warmup ratio 0.1, and max length 160. Only the enabled modalities differ. Checkpoints are selected by validation pair-role macro F1.

\begin{table}[t]
\centering
\caption{Architecture-specific training recipes.}
\label{tab:appendix-hparams}
\small
\setlength{\tabcolsep}{4pt}
\begin{tabular}{@{}>{\raggedright\arraybackslash}p{0.15\linewidth}>{\raggedright\arraybackslash}p{0.17\linewidth}>{\raggedright\arraybackslash}p{0.55\linewidth}@{}}
\toprule
Model & Optimizer & Distinct settings \\
\midrule
\mecpetwostep{} & Adam & step 1: batch 8, lr $5\times10^{-3}$, 15 iters; step 2: batch 64, lr $5\times10^{-3}$, 12 iters \\
\addlinespace[2pt]
HiLo & AdamW & BERT lr $1\times10^{-5}$, task lr $1\times10^{-4}$, batch 4, 20 epochs \\
\addlinespace[2pt]
\mthreehg{} & AdamW & train batch 8 with gradient accumulation 2; eval batch 8; lr $5\times10^{-5}$; 50 epochs \\
\bottomrule
\end{tabular}
\end{table}

Table~\ref{tab:appendix-hparams} isolates settings that vary for the stronger model families. The shared baseline recipe fixes the simple unimodal and fusion controls, while the table reports only architecture-specific choices for \mecpetwostep{}, HiLo, and \mthreehg{}. This separation helps interpret model-family differences without turning the appendix into a hyperparameter search.

\subsection{Model Size and Parameter Counts}
\label{app:model-size}

Table~\ref{tab:model_sizes} reports the parameter counts for the evaluated model families. For settings that use precomputed WavLM or CLIP features, we report the trainable predictor parameters and exclude fixed feature tables. MECPE-2step counts also exclude fixed embedding or feature tables. Fusion and graph-model rows report the instantiated model sizes for the corresponding modality settings.

\begin{table}[t]
\centering
\small
\caption{Model size summary. Counts are reported in millions of parameters. For feature-based audio/video settings and MECPE-2step, fixed feature or embedding tables are excluded as noted in the text.}
\label{tab:model_sizes}
\setlength{\tabcolsep}{2pt}
\resizebox{\linewidth}{!}{%
\begin{tabular}{ll}
\toprule
\textbf{Model setting} & \textbf{Parameters} \\
\midrule
RoBERTa (T) & 124.65M \\
WavLM feature MLP (A) & 0.39M trainable \\
CLIP feature MLP (V) & 0.39M trainable \\
RW / RC / RWC-Fusion & 125.83M / 125.83M / 127.02M \\
MECPE-2step, T to T+A+V & 1.82M--2.77M trainable \\
HiLo (T+A+V) & 153.28M \\
\mthreehg{}, T to T+A+V & 171.61M--181.85M \\
\bottomrule
\end{tabular}
}
\end{table}

\subsection{Compute Environment}
\label{app:compute-environment}

The reported experiments were run on the server summarized in Table~\ref{tab:appendix-environment}. The core multimodal pipeline uses the Python/PyTorch stack listed there, while \mecpetwostep{} is executed through a separate TensorFlow-compatible launcher.

\begin{table}[t]
\centering
\caption{Compute environment and key package versions.}
\label{tab:appendix-environment}
\small
\setlength{\tabcolsep}{3pt}
\begin{tabular}{@{}>{\raggedright\arraybackslash}p{0.30\linewidth}>{\raggedright\arraybackslash}p{0.60\linewidth}@{}}
\toprule
Item & Value \\
\midrule
Operating system & Linux 5.15.0-1045-nvidia \\
GPUs & 2 $\times$ NVIDIA A100-SXM4-80GB \\
Driver & 535.216.03 \\
Python & 3.12.12 \\
PyTorch & 2.10.0+cu128 \\
transformers & 5.7.0 \\
numpy & 1.26.4 \\
pandas & 3.0.2 \\
matplotlib & 3.10.9 \\
scikit-learn & 1.8.0 \\
tqdm & 4.67.3 \\
PyYAML & 6.0.3 \\
seaborn & 0.13.2 \\
MECPE-2step launcher & separate TensorFlow 2.15-compatible environment \\
\bottomrule
\end{tabular}
\end{table}

The experimental budget consists of 54 seeded three-class pair-role runs, covering 18 model-modality settings over three seeds, and 18 seeded binary-source diagnostic runs. This yields 72 one-GPU training/evaluation runs. Each run used one A100 80GB GPU. Summing the per-job durations recoverable from launcher logs gives approximately 65.3 A100 GPU-hours, equivalent to about 32.6 fully utilized hours on the two-A100 server.

\subsection{AI-Assistance Disclosure}
\label{app:ai-assistance}

OpenAI Codex with the model and reasoning setting reported as \texttt{gpt-5.5 xhigh} was used for LLM-assisted construction support. The outputs were used as proposal, review, and verification signals under the safeguards described in Section~\ref{sec:data-protocol}; they were not treated as unchecked gold annotations or unverified scientific claims. The recorded Codex usage for the annotation and dataset-construction workflow totaled approximately 509 million tokens. This disclosure is included because \dataset{} is positioned as a diagnostic audit rather than as a fully human-reannotated corpus from scratch.

\begin{figure}[t]
  \centering
  \includegraphics[width=0.9\linewidth]{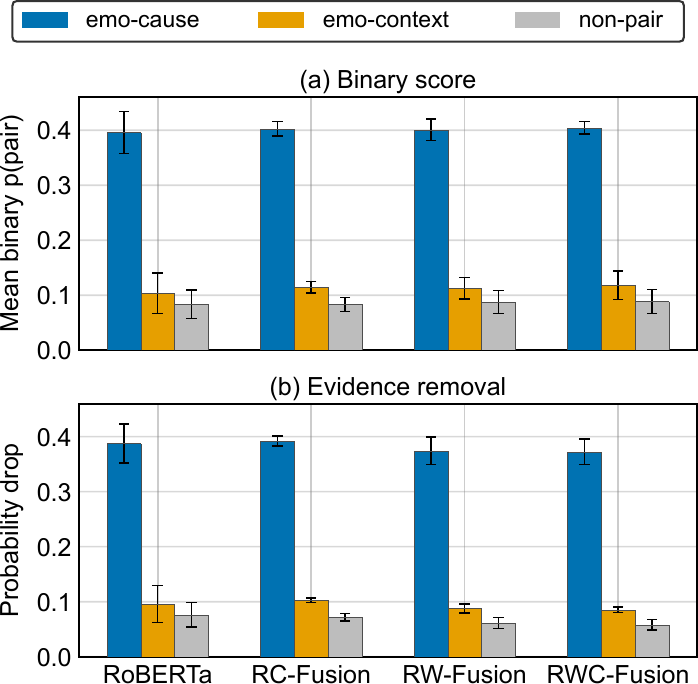}
  \caption{Binary-axis and evidence-removal diagnostics across text-grounded and fusion controls. Binary-source scores keep \context{} closer to \nonpair{} than to \cause{}.}
  \label{fig:appendix-binary-axis-removal-family}
\end{figure}

\begin{figure*}[t]
  \centering
  \includegraphics[width=0.95\textwidth]{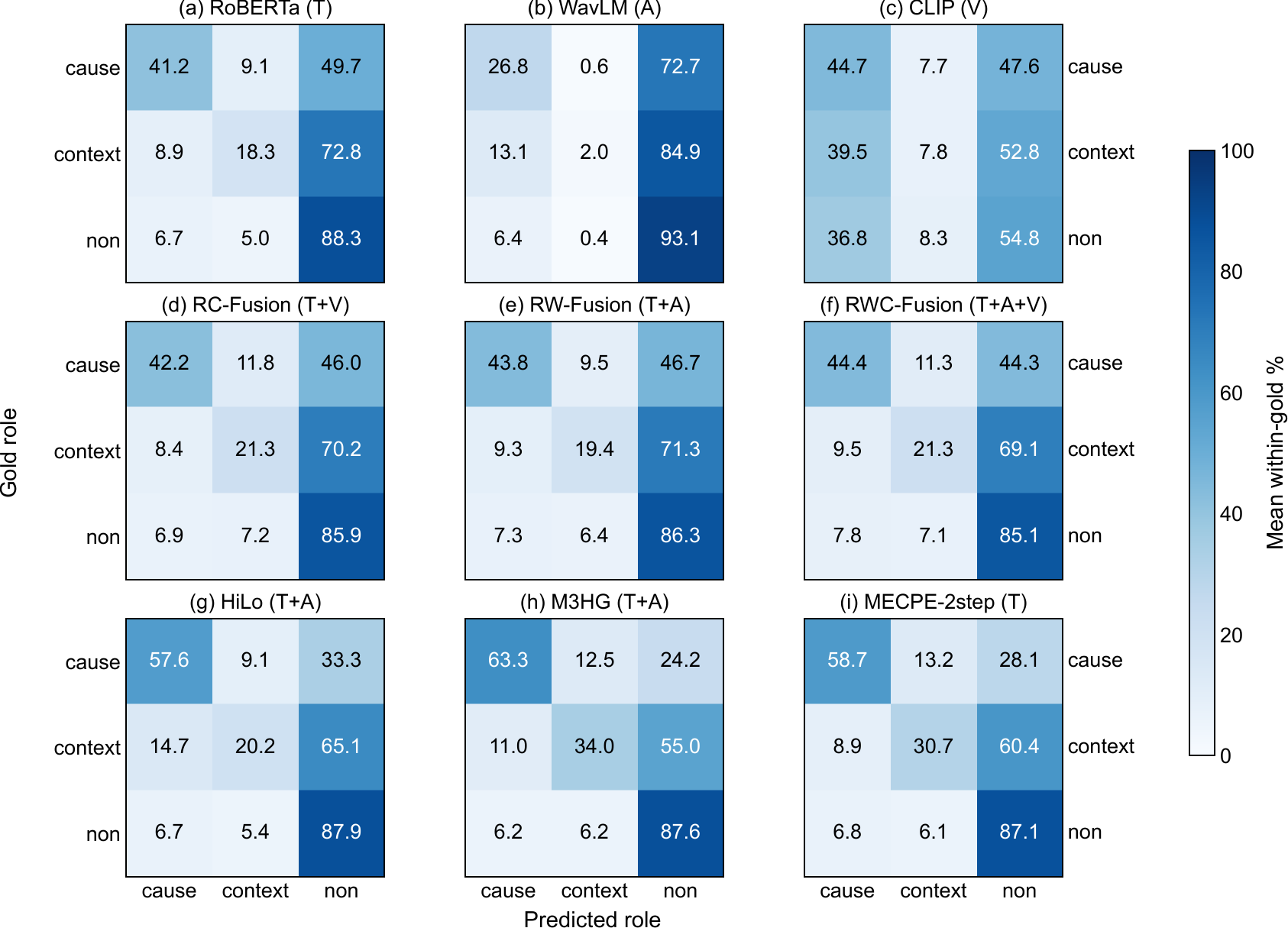}
  \caption{Row-normalized pair-role confusion matrices under three-class supervision. Across model families, the dominant \context{} error is collapse into \nonpair{}.}
  \label{fig:appendix-family-confusion}
  \vspace{-1em}
\end{figure*}

\section{Boundary Evidence and Cross-Family Diagnostics}
\label{app:boundary-evidence}

This section expands the model-side diagnostics behind the main text. The central question is whether the same validity pattern persists across model families and controls. It is found that binary ECPE preserves useful direct-cause decisions, but its score geometry and error patterns remain weak for non-triggering discourse evidence.

\subsection{Binary-Axis and Evidence-Removal Diagnostics}
\label{app:binary-axis-removal-family}

Figure~\ref{fig:appendix-binary-axis-removal-family} repeats the binary-axis and candidate-removal diagnostics across text-grounded and fusion controls. The binary-axis component asks where final \cause{}, \context{}, and \nonpair{} pairs fall under original binary supervision. The candidate-removal component asks how much the relevant score drops when the candidate utterance is removed.

The finding matches the main RWC-Fusion diagnostic. Gold \context{} receives binary pair scores closer to \nonpair{} than to \cause{}, which indicates that binary supervision lacks a stable context channel. Candidate removal also affects \cause{} more strongly than \context{} under binary-source scoring. The model-side signal is therefore not simply that \context{} is difficult; it is that binary pair scores are more sensitive to direct-cause evidence than to non-triggering support.

\subsection{Cross-Family Confusion Matrices}
\label{app:family-confusion}

Figure~\ref{fig:appendix-family-confusion} shows mean 3$\times$3 pair-role confusion matrices for the evaluated model families under three-class supervision. Cells are row-normalized, so each row describes where a gold role is predicted rather than how frequent that role is in the dataset.

The confusion matrices support the role-wise conclusion in the main text. Context collapse is not confined to one baseline or one modality setting. Across model families, gold \context{} is frequently absorbed into \nonpair{}, even when the same systems identify many \cause{} and \nonpair{} examples more reliably. This makes the error direction part of the diagnostic evidence. Models do not merely miss a hard class; they tend to treat contextual support as absence of pair evidence.

\begin{table*}[!t]
  \centering
  \caption{Natural shortcut-vs-evidence stress-test results for binary-source controls. Values are mean$_{\pm\mathrm{std}}$ over seeds; $n$, distance, and lexical overlap are fixed by subset definition.}
  \label{tab:appendix-shortcut-vs-evidence}
  \small
  \begin{tabular}{llrrrrr}
    \toprule
    Model & Stress subset & $n$ & $p(\mathrm{pair})$ & Pred-pair (\%) & Dist. & Lex. overlap \\
    \midrule
    \multirow{3}{*}{RoBERTa}
      & Local lexical \nonpair{} & 1489 & 0.302$_{\pm0.041}$ & 29.0$_{\pm3.5}$ & 0.3 & 0.78 \\
      & Long-distance \cause{} & 92 & 0.139$_{\pm0.033}$ & 11.6$_{\pm1.7}$ & 9.6 & 0.07 \\
      & Original-negative \context{} & 854 & 0.105$_{\pm0.037}$ & 7.5$_{\pm1.1}$ & 4.5 & 0.07 \\
    \midrule
    \multirow{3}{*}{RC-Fusion}
      & Local lexical \nonpair{} & 1489 & 0.302$_{\pm0.012}$ & 28.9$_{\pm2.6}$ & 0.3 & 0.78 \\
      & Long-distance \cause{} & 92 & 0.143$_{\pm0.028}$ & 11.6$_{\pm2.7}$ & 9.6 & 0.07 \\
      & Original-negative \context{} & 854 & 0.117$_{\pm0.011}$ & 9.4$_{\pm2.3}$ & 4.5 & 0.07 \\
    \midrule
    \multirow{3}{*}{RW-Fusion}
      & Local lexical \nonpair{} & 1489 & 0.292$_{\pm0.018}$ & 28.3$_{\pm1.4}$ & 0.3 & 0.78 \\
      & Long-distance \cause{} & 92 & 0.147$_{\pm0.020}$ & 10.5$_{\pm1.7}$ & 9.6 & 0.07 \\
      & Original-negative \context{} & 854 & 0.115$_{\pm0.019}$ & 7.2$_{\pm1.6}$ & 4.5 & 0.07 \\
    \midrule
    \multirow{3}{*}{RWC-Fusion}
      & Local lexical \nonpair{} & 1489 & 0.300$_{\pm0.023}$ & 29.4$_{\pm2.5}$ & 0.3 & 0.78 \\
      & Long-distance \cause{} & 92 & 0.147$_{\pm0.037}$ & 11.6$_{\pm4.1}$ & 9.6 & 0.07 \\
      & Original-negative \context{} & 854 & 0.119$_{\pm0.024}$ & 8.7$_{\pm1.6}$ & 4.5 & 0.07 \\
    \bottomrule
  \end{tabular}
\end{table*}

\subsection{Full Shortcut-vs-Evidence Stress Test}
\label{app:shortcut-vs-evidence}

Table~\ref{tab:appendix-shortcut-vs-evidence} expands the natural shortcut-vs-evidence stress test from Section~\ref{sec:shortcut-main}. The local lexical \nonpair{} subset is not adversarially edited. It consists of naturally occurring candidates that are close to the target and lexically overlapping with it, but unsupported under the final role label. The evidence-supported subsets have weaker shortcut alignment, where long-distance \cause{} pairs are structurally non-local, and original-negative \context{} pairs come from the inherited negative side.

The ordering is stable across binary-source text-grounded and fusion families. Local lexical \nonpair{} controls receive higher mean pair scores and higher predicted-pair rates than both evidence-supported subsets. The distance and lexical-overlap columns explain why this is a shortcut-conflict test. The unsupported control is close and overlapping, whereas the evidence-supported subsets are farther and less lexically aligned. The table therefore supports the main claim that binary pair scores can be inflated by shortcut-compatible structure.

\begin{table*}[!t]
  \centering
  \caption{Proxy-only prediction of trained-model behavior. Values are mean$_{\pm\mathrm{std}}$ over seeds. AUC and accuracy are reported for hard predictions; $R^2$ and Spearman $\rho$ are reported for score regression. Dashes mark metrics not defined for the target type.}
  \label{tab:appendix-proxy-only}
  \small
  \begin{tabular}{llllrrrr}
    \toprule
    Source & Target & Features & Model & AUC & Acc. & $R^2$ & $\rho$ \\
    \midrule
    \multirow{8}{*}{Binary}
      & \multirow{4}{*}{hard pair}
      & \multirow{4}{*}{Obs.}
      & RoBERTa & 0.855$_{\pm0.007}$ & 79.6$_{\pm1.1}$ & - & - \\
      & & & RC-Fusion & 0.850$_{\pm0.020}$ & 79.1$_{\pm2.0}$ & - & - \\
      & & & RW-Fusion & 0.851$_{\pm0.025}$ & 80.1$_{\pm2.7}$ & - & - \\
      & & & RWC-Fusion & 0.833$_{\pm0.028}$ & 78.8$_{\pm2.3}$ & - & - \\
    \cmidrule(lr){2-8}
      & \multirow{4}{*}{hard pair}
      & \multirow{4}{*}{Obs.+src}
      & RoBERTa & 0.858$_{\pm0.006}$ & 81.0$_{\pm1.0}$ & - & - \\
      & & & RC-Fusion & 0.853$_{\pm0.021}$ & 80.2$_{\pm2.1}$ & - & - \\
      & & & RW-Fusion & 0.854$_{\pm0.025}$ & 81.4$_{\pm2.7}$ & - & - \\
      & & & RWC-Fusion & 0.836$_{\pm0.029}$ & 80.2$_{\pm2.2}$ & - & - \\
    \midrule
    \multirow{8}{*}{Binary}
      & \multirow{4}{*}{$p(\mathrm{pair})$}
      & \multirow{4}{*}{Obs.}
      & RoBERTa & - & - & 0.282$_{\pm0.058}$ & 0.548$_{\pm0.063}$ \\
      & & & RC-Fusion & - & - & 0.269$_{\pm0.030}$ & 0.462$_{\pm0.114}$ \\
      & & & RW-Fusion & - & - & 0.295$_{\pm0.058}$ & 0.605$_{\pm0.054}$ \\
      & & & RWC-Fusion & - & - & 0.263$_{\pm0.079}$ & 0.562$_{\pm0.085}$ \\
    \cmidrule(lr){2-8}
      & \multirow{4}{*}{$p(\mathrm{pair})$}
      & \multirow{4}{*}{Obs.+src}
      & RoBERTa & - & - & 0.304$_{\pm0.057}$ & 0.555$_{\pm0.062}$ \\
      & & & RC-Fusion & - & - & 0.293$_{\pm0.031}$ & 0.472$_{\pm0.116}$ \\
      & & & RW-Fusion & - & - & 0.321$_{\pm0.059}$ & 0.615$_{\pm0.056}$ \\
      & & & RWC-Fusion & - & - & 0.286$_{\pm0.081}$ & 0.571$_{\pm0.086}$ \\
    \midrule
    \multirow{8}{*}{Three-class}
      & \multirow{4}{*}{hard context}
      & \multirow{4}{*}{Obs.}
      & RoBERTa & 0.855$_{\pm0.061}$ & 77.2$_{\pm5.6}$ & - & - \\
      & & & RC-Fusion & 0.830$_{\pm0.051}$ & 74.7$_{\pm4.7}$ & - & - \\
      & & & RW-Fusion & 0.844$_{\pm0.043}$ & 75.7$_{\pm3.0}$ & - & - \\
      & & & RWC-Fusion & 0.833$_{\pm0.057}$ & 74.9$_{\pm5.0}$ & - & - \\
    \cmidrule(lr){2-8}
      & \multirow{4}{*}{hard context}
      & \multirow{4}{*}{Obs.+src}
      & RoBERTa & 0.855$_{\pm0.061}$ & 77.2$_{\pm5.6}$ & - & - \\
      & & & RC-Fusion & 0.830$_{\pm0.051}$ & 74.7$_{\pm4.6}$ & - & - \\
      & & & RW-Fusion & 0.843$_{\pm0.043}$ & 75.7$_{\pm3.0}$ & - & - \\
      & & & RWC-Fusion & 0.832$_{\pm0.057}$ & 74.9$_{\pm5.1}$ & - & - \\
    \midrule
    \multirow{8}{*}{Three-class}
      & \multirow{4}{*}{$p(\mathrm{context})$}
      & \multirow{4}{*}{Obs.}
      & RoBERTa & - & - & 0.244$_{\pm0.153}$ & 0.537$_{\pm0.154}$ \\
      & & & RC-Fusion & - & - & 0.229$_{\pm0.124}$ & 0.486$_{\pm0.121}$ \\
      & & & RW-Fusion & - & - & 0.219$_{\pm0.135}$ & 0.510$_{\pm0.134}$ \\
      & & & RWC-Fusion & - & - & 0.258$_{\pm0.163}$ & 0.547$_{\pm0.137}$ \\
    \cmidrule(lr){2-8}
      & \multirow{4}{*}{$p(\mathrm{context})$}
      & \multirow{4}{*}{Obs.+src}
      & RoBERTa & - & - & 0.243$_{\pm0.153}$ & 0.537$_{\pm0.154}$ \\
      & & & RC-Fusion & - & - & 0.229$_{\pm0.125}$ & 0.486$_{\pm0.122}$ \\
      & & & RW-Fusion & - & - & 0.218$_{\pm0.135}$ & 0.510$_{\pm0.134}$ \\
      & & & RWC-Fusion & - & - & 0.258$_{\pm0.163}$ & 0.546$_{\pm0.137}$ \\
    \bottomrule
  \end{tabular}
\end{table*}

\subsection{Full Proxy-Only Behavior Prediction}
\label{app:proxy-only}

Table~\ref{tab:appendix-proxy-only} reports the broader proxy-only probe. Observable-only features include structural, dialogue, speaker, emotion, position, length, and lexical-overlap variables, but not utterance text or multimodal content. The source-binary variant adds inherited binary metadata. The probe predicts trained-model outputs, not gold labels. High values therefore indicate that model decision or score behavior is recoverable from shortcut-compatible observable structure.

The binary-source rows show that observable-only features recover hard pair decisions with high AUC and recover pair-score rankings with moderate to strong correlations. The three-class rows show a related issue for context behavior, where context decisions and context probabilities are also partly predictable from structural variables. This does not imply that the ECPE models ignore utterance evidence. It shows that score behavior is substantially aligned with shortcut-compatible structure, which is why binary pair scores should not be exposed as faithful evidence-grounded explanations without role-level diagnostics.

\begin{table*}[t]
  \centering
  \caption{Calibration baselines for role-wise evaluation on the test split. F1 values are percentages. The \mthreehg{} T+A row is the main text reference row.}
  \label{tab:appendix-calibration-baselines}
  \small
  \begin{tabular}{lrrrr}
    \toprule
    Method & Macro F1 & \cause{} F1 & \context{} F1 & \nonpair{} F1 \\
    \midrule
    Distance-bucket majority (= always \nonpair{}) & 30.58 & 0.00 & 0.00 & 91.73 \\
    Distance-only balanced logistic regression & 48.31 & 47.40 & 21.99 & 75.53 \\
    Binary-source predictor $\rightarrow$ pair-role remap & 43.35 & 39.32 & 0.00 & 90.71 \\
    \midrule
    \mthreehg{} T+A & 57.95 & 56.73 & 26.84 & 90.30 \\
    \bottomrule
  \end{tabular}
\end{table*}

\begin{table*}[!t]
  \centering
  \caption{Source-binary Pair F1 after remapping three-class predictions. Values are mean$_{\pm\mathrm{std}}$ over seeds 42, 123, and 456.}
  \label{tab:appendix-source-binary-remap}
  \small
  \begin{tabular}{llrrrr}
    \toprule
    Model & Mod. & Original binary & Drop \context{} & \context{} $\rightarrow$ pair & \context{} $\rightarrow$ \nonpair{} \\
    \midrule
    RoBERTa & T & 37.60$_{\pm1.05}$ & 40.87$_{\pm0.34}$ & 36.70$_{\pm0.48}$ & 39.05$_{\pm0.06}$ \\
    WavLM & A & 28.24$_{\pm0.22}$ & 27.67$_{\pm0.24}$ & 27.48$_{\pm0.30}$ & 27.58$_{\pm0.23}$ \\
    CLIP & V & 16.43$_{\pm0.94}$ & 19.06$_{\pm0.62}$ & 18.71$_{\pm0.15}$ & 18.71$_{\pm0.33}$ \\
    RW-Fusion & T+A & 37.86$_{\pm0.68}$ & 41.95$_{\pm1.18}$ & 36.06$_{\pm1.34}$ & 40.12$_{\pm0.98}$ \\
    RC-Fusion & T+V & 37.07$_{\pm0.28}$ & 42.20$_{\pm0.69}$ & 36.25$_{\pm1.92}$ & 39.88$_{\pm0.99}$ \\
    RWC-Fusion & T+A+V & 37.61$_{\pm0.22}$ & 41.97$_{\pm1.33}$ & 36.11$_{\pm0.21}$ & 39.78$_{\pm0.86}$ \\
    \bottomrule
  \end{tabular}
\end{table*}

\begin{table*}[!t]
  \centering
  \caption{Full family-level test results over three seeds.}
  \label{tab:appendix-full-results}
  \small
  \begin{tabular}{llrrrrr}
    \toprule
    Family & Mod. & Macro F1 & Acc. & \cause{} F1 & \context{} F1 & \nonpair{} F1 \\
    \midrule
    RoBERTa & T & 48.77$_{\pm0.39}$ & 79.83$_{\pm0.41}$ & 40.28$_{\pm0.52}$ & 17.18$_{\pm1.47}$ & 88.84$_{\pm0.29}$ \\
    WavLM & A & 40.63$_{\pm0.33}$ & 81.64$_{\pm0.15}$ & 28.26$_{\pm0.23}$ & 3.65$_{\pm0.73}$ & 89.97$_{\pm0.10}$ \\
    CLIP & V & 30.19$_{\pm1.32}$ & 51.26$_{\pm3.61}$ & 18.50$_{\pm0.22}$ & 5.20$_{\pm2.02}$ & 66.86$_{\pm3.36}$ \\
    RW-Fusion & T+A & 48.54$_{\pm0.26}$ & 78.47$_{\pm0.57}$ & 41.37$_{\pm1.15}$ & 16.34$_{\pm1.70}$ & 87.92$_{\pm0.37}$ \\
    RC-Fusion & T+V & 48.38$_{\pm0.63}$ & 78.12$_{\pm1.82}$ & 40.98$_{\pm1.07}$ & 16.40$_{\pm2.80}$ & 87.76$_{\pm1.23}$ \\
    RWC-Fusion & T+A+V & 48.41$_{\pm0.50}$ & 77.61$_{\pm0.68}$ & 40.89$_{\pm0.63}$ & 16.94$_{\pm1.74}$ & 87.41$_{\pm0.52}$ \\
    \midrule
    \multirow{4}{*}{\mecpetwosteptable}
      & T & 55.73$_{\pm0.34}$ & 81.26$_{\pm0.25}$ & 52.91$_{\pm0.44}$ & 24.60$_{\pm0.82}$ & 89.69$_{\pm0.14}$ \\
      & T+A & 55.60$_{\pm0.08}$ & 79.67$_{\pm0.94}$ & 52.98$_{\pm0.11}$ & 25.14$_{\pm0.91}$ & 88.67$_{\pm0.60}$ \\
      & T+A+V & 55.24$_{\pm0.38}$ & 79.73$_{\pm0.66}$ & 52.55$_{\pm0.49}$ & 24.51$_{\pm1.11}$ & 88.65$_{\pm0.42}$ \\
      & T+V & 55.46$_{\pm0.11}$ & 80.15$_{\pm0.35}$ & 53.02$_{\pm0.31}$ & 24.41$_{\pm0.63}$ & 88.95$_{\pm0.20}$ \\
    \midrule
    \multirow{4}{*}{HiLo}
      & T & 52.50$_{\pm0.21}$ & 80.51$_{\pm0.80}$ & 50.53$_{\pm0.95}$ & 17.73$_{\pm0.76}$ & 89.25$_{\pm0.53}$ \\
      & T+A & 52.98$_{\pm0.80}$ & 81.25$_{\pm1.32}$ & 51.29$_{\pm0.44}$ & 17.95$_{\pm2.67}$ & 89.72$_{\pm0.86}$ \\
      & T+A+V & 52.94$_{\pm0.61}$ & 79.66$_{\pm1.55}$ & 50.60$_{\pm0.69}$ & 19.60$_{\pm2.06}$ & 88.61$_{\pm0.99}$ \\
      & T+V & 51.85$_{\pm0.62}$ & 82.10$_{\pm1.11}$ & 48.59$_{\pm0.61}$ & 16.69$_{\pm1.64}$ & 90.26$_{\pm0.63}$ \\
    \midrule
    \multirow{4}{*}{\mthreehg{}}
      & T & 57.84$_{\pm0.11}$ & 80.88$_{\pm0.52}$ & 56.34$_{\pm0.41}$ & 27.73$_{\pm0.57}$ & 89.46$_{\pm0.36}$ \\
      & T+A & 57.95$_{\pm0.07}$ & 82.24$_{\pm0.91}$ & 56.73$_{\pm0.18}$ & 26.84$_{\pm0.33}$ & 90.30$_{\pm0.52}$ \\
      & T+A+V & 57.76$_{\pm0.29}$ & 82.66$_{\pm0.18}$ & 55.52$_{\pm0.07}$ & 27.10$_{\pm1.07}$ & 90.67$_{\pm0.14}$ \\
      & T+V & 57.18$_{\pm0.32}$ & 81.75$_{\pm0.14}$ & 55.55$_{\pm0.73}$ & 25.88$_{\pm1.45}$ & 90.11$_{\pm0.11}$ \\
    \bottomrule
  \end{tabular}
\end{table*}

\section{Full Family-Level Results}
\label{app:full-results}

This section preserves the full model evidence behind the compact main text results.

\subsection{Calibration Baselines for Role-Wise Results}
\label{app:calibration-baselines}

Table~\ref{tab:appendix-calibration-baselines} provides simple calibration baselines for the role-wise results in Table~\ref{tab:diagnostic-main-results}. These baselines are diagnostic rather than competitive. They clarify whether the three-class pair-role results can be explained by majority-class behavior, distance structure, or the inherited binary output space. The distance-bucket majority baseline collapses to always-\nonpair{} behavior, and its high \nonpair{} F1 but zero \cause{} and \context{} F1 confirms why macro and role-wise F1 are more informative than majority-class behavior.

The distance-only balanced logistic regression obtains non-zero \context{} F1, showing that \context{} contains learnable distance regularity rather than behaving as random noise. However, the binary-source predictor remapped into the pair-role space still has no \context{} channel, because predicted pairs can only be projected to \cause{} and predicted non-pairs to \nonpair{}. The \mthreehg{} T+A reference setting gives the strongest macro F1 among these rows while preserving a non-zero \context{} channel and high \nonpair{} F1. These calibration results support the main text interpretation, which is that the three-class pair-role results are not explained by majority behavior or a simple distance heuristic, and the inherited binary output space cannot express non-triggering contextual support as a separate role.

\subsection{Source-Binary Remapping}
\label{app:source-binary-remap}

Table~\ref{tab:appendix-source-binary-remap} maps three-class predictions back to the inherited binary ECPE task. ``Drop \context{}'' evaluates only non-context predictions under the source-binary mapping. ``\context{} $\rightarrow$ pair'' treats predicted \context{} as source-positive. ``\context{} $\rightarrow$ \nonpair{}'' treats predicted \context{} as source-negative.

The remapping results support the compatibility claim in the main text. Treating predicted \context{} as source-positive generally hurts text-grounded settings, so the role should not be collapsed into a simple positive expansion. In contrast, dropping predicted \context{} or mapping it to \nonpair{} preserves or improves source-binary Pair F1 for the text-grounded and fusion controls. Separating \context{} can therefore sharpen the direct-cause channel without redefining every contextual support case as a cause.

\subsection{Full Model-Modality Results}
\label{app:full-family-results}

Table~\ref{tab:appendix-full-results} reports mean and standard deviation over three seeds for all 18 model-modality settings. T, A, and V denote text, audio, and video inputs. Rows grouped under a family name share the same architecture with different modality choices.

The full table preserves the role-wise pattern behind the compact main text table. The highest macro F1 among these rows is achieved by \mthreehg{} T+A, while the highest \context{} F1 is achieved by \mthreehg{} T. This separation reinforces why aggregate scores should not be read as sufficient evidence of explanation quality. Accuracy and \nonpair{} F1 are high partly because the candidate space is dominated by \nonpair{}, whereas \context{} F1 remains substantially lower than \cause{} and \nonpair{} F1 across families. The full results therefore support the same conclusion as the main text, where current systems improve direct-cause extraction more reliably than contextual support.

\section{Annotation Instructions and Visibility}
\label{app:prompts}

This section reports the operational instructions used for the Codex-assisted draft pass, targeted human refinement, and blind human re-annotation audit. The first two stages are not blind: both had access to the inherited source-binary label as audit metadata, because they target boundary cases in which contextual support may be over-counted as pair or discarded as non-pair. In contrast, the blind re-annotation audit removes construction metadata; annotators see only dialogue, target, and candidate information.

\subsection{Codex-Assisted Draft Pass}
\label{app:codex-context-mining-prompt}

Table~\ref{tab:codex-context-mining-prompt} gives the instruction used for the Codex-assisted draft pass. Codex produces a pair-role proposal and a short evidence-grounded reason for each reviewed candidate. The source-binary label is used only to identify the boundary being audited, not as a gold role label.

\begin{table*}[p]
	\centering
	\caption{Instruction used for the Codex-assisted draft pass over inherited ECPE candidate pairs.}
	\label{tab:codex-context-mining-prompt}
	\begin{promptboxside}{Prompt: Codex-Assisted Draft Pass}
		You are auditing ECPE candidate pairs.
		
		Goal:
		Propose one pair role for each reviewed candidate and identify contextual support that may be compressed by the source-binary pair/non-pair label space.
		
		You may see the source-binary label. Use it as audit metadata only:
		- source pair: the candidate may be a direct cause, or it may be context that was over-counted as causal.
		- source non-pair: the candidate may be unsupported, or it may be contextual support missed by the binary label.
		Do not treat the source-binary label as the final pair role.
		
		Available information:
		- dialogue identifier and split
		- target turn, target speaker, target emotion, target text
		- candidate turn, candidate speaker, candidate text
		- source-binary label or current draft role, when available
		- dialogue context
		
		Choose one pair role:
		
		emo-cause:
		The candidate directly triggers, explains, states, realizes, or appraises the target emotion. A self-pair is emo-cause only when the target utterance itself contains the event, reason, appraisal, or desire-obstacle conflict.
		
		emo-context:
		The candidate is not the direct trigger, but it provides interpretive support for the target emotion. It may provide background, conflict setup, relationship state, discourse bridge, prior speaker state, situational constraint, or an alternative plausible cause. Use emo-context only when removing the candidate would weaken interpretation of the target emotion.
		
		non-pair:
		The candidate does not provide direct-cause evidence or contextual support. Same-topic continuation, nearby position, speaker overlap, lexical overlap, or vague mood match is not enough.
		
		Priority rules:
		1. Direct cause beats context.
		2. Context is not a weak-cause or uncertain-cause label.
		3. For source pair -> emo-context, explain why the candidate supports interpretation but does not directly trigger the target emotion.
		4. For source non-pair -> emo-context, explain what discourse evidence the binary label missed.
		5. If evidence is ambiguous or only topical, choose non-pair.
		
		Output one reviewed row per candidate with:
		- pair identifiers
		- source-binary label or current draft role, when available
		- proposed pair role
		- short evidence-grounded reason
	\end{promptboxside}
\end{table*}

\subsection{Targeted Human Refinement Rules}
\label{app:prompt-refinement-section}

Table~\ref{tab:targeted-human-refinement} reports the rules used for targeted human refinement. This stage checks Codex-proposed labels and boundary-sensitive rows; it is not an independent blind relabeling of the full candidate space. Reviewers may inspect the source-binary label, the Codex proposal, and the dialogue context in order to decide whether the proposed boundary movement is valid.

The rules are conservative about \context{}. A reviewer confirms \context{} only when the candidate supplies observable, non-triggering evidence that makes the target emotion easier to interpret. Otherwise, the reviewer keeps or restores \cause{} for direct triggers and \nonpair{} for unsupported candidates.

\begin{table*}[p]
	\centering
	\caption{Instruction used for targeted human refinement of Codex-proposed pair roles.}
	\label{tab:targeted-human-refinement}
	\begin{promptboxside}{Prompt: Targeted Human Refinement}
		You are validating targeted Codex pair-role proposals.
		
		This is not a blind task. You may see:
		- source-binary label
		- Codex proposed role and rationale
		- transition type
		- target and candidate text
		- full dialogue context
		
		Task:
		Approve, reject, or correct the proposed pair role by assigning one final pair role.
		
		Allowed final pair roles:
		- emo-cause
		- emo-context
		- non-pair
		
		Rules:
		1. Treat the Codex label as a proposal, not as authority.
		2. Treat the source-binary label as boundary metadata, not as authority.
		3. Use emo-cause only for observable direct-cause evidence.
		4. Use emo-context only for non-triggering discourse evidence that helps interpret the target emotion.
		5. Keep a source pair as emo-cause when it directly explains the target; change it to emo-context only when it is explanatory background rather than the trigger.
		6. Change a source non-pair to emo-context only when the candidate provides concrete interpretive support missed by the binary label.
		7. Same-topic relation, adjacency, lexical overlap, or same speaker is insufficient for emo-context.
		8. For self-pairs, use emo-cause only when the target utterance contains its own event, reason, appraisal, or desire-obstacle conflict.
		9. When evidence is too weak or ambiguous to support cause or context, choose non-pair.
		
		Output:
		- final pair role
		- short evidence-grounded reason
	\end{promptboxside}
\end{table*}

\subsection{Blind Human Re-Annotation Audit Instructions}
\label{app:blind-reannotation-instructions}

The blind re-annotation audit uses the same role definitions but removes construction metadata from the annotator view. Table~\ref{tab:blind-reannotation-instructions} gives the instruction. This separation matters because Codex-assisted drafting and targeted human refinement are construction and verification steps, while the blind audit measures whether the pair-role boundary can be recovered without seeing source labels, proposals, final labels, provenance, or sampling strata.

\begin{table*}[p]
	\centering
	\caption{Instruction used for the blind human re-annotation audit.}
	\label{tab:blind-reannotation-instructions}
	\begin{promptboxside}{Prompt: Blind Pair-Role Re-Annotation}
		You are independently re-annotating candidate-to-target emotion pairs.
		This is a blind audit.
		
		You may see:
		- dialogue context
		- target turn, target speaker, target emotion, target text
		- candidate turn, candidate distance, candidate speaker,
		candidate emotion, candidate text
		
		You must not see or use:
		- source-binary pair/non-pair label
		- Codex proposed role or rationale
		- final pair-role label
		- label source or provenance
		- packet type, sampling stratum, or boundary category
		- other annotators' judgments
		
		For each candidate, choose one pair role:
		
		emo-cause:
		The candidate directly triggers, explains, states, realizes, or appraises the target emotion.
		
		emo-context:
		The candidate is not the direct trigger, but it provides background, setup, relationship state, prior state, discourse bridge, or situational information needed to interpret the target emotion.
		
		non-pair:
		The candidate does not provide enough evidence for direct causality or contextual support.
		
		Decision rules:
		1. Direct cause beats context.
		2. Context must be evidence-bearing; proximity or topical similarity is not enough.
		3. If the candidate only makes the dialogue generally coherent but does not help interpret the target emotion, choose non-pair.
		4. If unsure between cause and context, choose the best role and mark causal-chain ambiguity.
		5. If unsure because evidence is absent or too weak, choose non-pair.
		
		Output:
		- pair role: emo-cause | emo-context | non-pair
		- confidence: high | medium | low
		- causal-chain ambiguity: yes | no
		- optional note for difficult cases
	\end{promptboxside}
\end{table*}

\section{Responsible Research Statement}
\label{app:responsible_research}

\subsection{Intended Use and Artifact Release}
\label{app:intended_use_release}

\dataset{} is intended as a research diagnostic artifact for studying label-space compression in ECPE. It supports dataset auditing, benchmark analysis, and evaluation diagnostics, and is not intended for deployed user-facing emotion explanation, clinical or legal decision making, employment decisions, educational assessment, or other individual-level decisions.

The artifact builds on IEMOCAP-derived ConvECPE resources \citep{busso2008iemocap,li2023ecpec} and preserves the inherited candidate-pair structure while adding pair-role labels and diagnostic metadata. We will release the \dataset{} pair-role labels, split files, candidate-pair identifiers, diagnostic metadata, and code/scripts. We will not redistribute restricted raw audio, video, dialogue content, or other underlying data not created by us.

\subsection{Licensing and Data Protection}
\label{app:licensing_data_protection}

The pair-role labels, split files, and diagnostic metadata created by us will be released under the Creative Commons Attribution 4.0 International license (CC BY 4.0). The accompanying code and scripts will be released under the MIT License. Candidate-pair identifiers are included only to align the released labels with the inherited candidate-pair structure.

To reduce data-protection risks, the released files are limited to pair-role labels, candidate-pair identifiers, split information, and diagnostic metadata needed to reproduce the analyses. We do not add speaker names, demographic attributes, contact information, or other personally identifying fields. Before release, we check the annotation and metadata files to ensure that no new personally identifying fields are introduced.

\subsection{Human Annotation and Ethics Review}
\label{app:human_annotation_ethics}

Targeted human judging and the blind re-annotation audit were conducted by trained members of the author team, not by crowdworkers or external participants. The annotators were selected internally for their familiarity with NLP annotation, ECPE, and the pair-role definitions, and were not paid per item. The annotation work was part of normal research activity. The author-annotators were informed that their judgments would be used for research validation, dataset construction, and the released \dataset{} annotation files. We did not collect personal data from annotators beyond their annotation decisions.

\end{document}